\documentclass{article}
\usepackage{iclr2027_conference,times}

\usepackage[percent]{overpic}
\usepackage{threeparttable}
\usepackage{booktabs}
\usepackage{multirow}
\usepackage{paralist}
\usepackage{algorithm,algorithmic}
\usepackage{graphicx}
\usepackage{subcaption}
\usepackage{times}
\usepackage{hyperref}
\usepackage{afterpage}

\usepackage{makecell}
\usepackage{pifont}
\usepackage{arydshln}
\usepackage[table]{xcolor}
\usepackage{mathtools}
\usepackage{url}
\usepackage{wrapfig}

\usepackage{amsmath,amssymb,amsfonts,bm,xspace}

\newenvironment{proof} {\par\noindent\textit{Proof.}\ } {\hfill$\square$\par}
\newtheorem{theorem}{Theorem}
\newtheorem{lemma}{Lemma}
\newtheorem{corollary}{Corollary}
\newtheorem{assumption}{Assumption}

\def \D {\mathcal{D}}
\def \E {\mathbb{E}}
\def \R {\mathbb{R}}
\def \op {\mathbf{op}}
\newcommand{\msign}{\mathrm{msign}}
\newcommand{\std}[1]{\,\raisebox{0.15ex}{\scriptsize $\pm\,#1$}}
\newcommand{\dram}{\textsf{DRAM}\xspace}
\newcommand{\dramf}{\textsf{DRAM-F}\xspace}

\title{DRAM: Memory-Efficient Muon via Rank \\ Reallocation in Principal-Random Subspaces}

\author{
Haomin Bai\textsuperscript{\rm 1,2},~Dingzhi Yu\textsuperscript{\rm 1,2},~Xuling Zhao\textsuperscript{\rm 1,2},~Sifan Yang\textsuperscript{\rm 1,2},~Yibo Wang\textsuperscript{\rm 1,2},~Lijun Zhang\textsuperscript{\rm 1,2}\thanks{Lijun Zhang is the corresponding author.}\\
$^1$State Key Laboratory of Novel Software Technology, Nanjing University, Nanjing, China\\
$^2$School of Artificial Intelligence, Nanjing University, Nanjing, China\\
\texttt{\{baihm,yudz,yangsf,wangyb,zhanglj\}@lamda.nju.edu.cn}, \\
\texttt{231300088@smail.nju.edu.cn}
}

\iclrfinalcopy
\hypersetup{hidelinks,
  pdftitle={DRAM: Memory-Efficient Muon via Rank Reallocation in Principal-Random Subspaces},
  pdfauthor={Haomin Bai, Dingzhi Yu, Xuling Zhao, Sifan Yang, Yibo Wang, Lijun Zhang}}

\begin{document}
\maketitle

\begin{abstract}
Memory-efficient Muon methods aim to reduce storage overhead while preserving strong performance, yet existing approaches compromise either convergence guarantees or practical efficiency.
In this paper, we focus on subspace optimization for Muon, where the choice of subspace determines the gradient information retained for updating.
To preserve dominant information while providing the coverage needed for stochastic convergence guarantees, we construct a principal-random subspace that combines leading singular directions of the stochastic gradient with random directions sampled from their orthogonal complement.
Under a fixed rank budget, there is a trade-off between principal exploitation and complementary coverage, making rank allocation important for practical optimization.
Our key observation is that the preferred balance shifts from principal-heavy early in training to random-heavy later.
Motivated by this shift, we propose \dram, a memory-efficient Muon optimizer that reallocates principal and random ranks across stages while maintaining momentum and performing matrix-sign normalization entirely in the subspace.
Theoretically, we establish convergence guarantees for general rank-allocation schedules with nondecreasing coverage factors, obtaining rates of $O(T^{-1/4})$ in the stochastic setting and $O(T^{-1/2})$ in the deterministic setting, with fixed allocation as a special case.
Using a simple linear reallocation schedule, \dram achieves the best low-memory performance across three pre-training scales and nine fine-tuning settings.
\end{abstract}

\section{Introduction}
Muon~\citep{2024_muon_jordan} has emerged as an effective matrix-structured optimizer for large language model (LLM) training~\citep{2025_muon_scalable_llmtraining,2025_practical_muon_pretraining}, with adoption in frontier LLMs including Kimi K3~\citep{2026_KimiK3_report}, GLM-5~\citep{2026_GLM5_report}, DeepSeek-V4~\citep{2026_DSV4_report}, and Qwen3.8-Next~\citep{2026_Qwen38_next_report}.
For each matrix parameter, Muon maintains a dense momentum that aggregates historical gradients and applies matrix-sign normalization before the parameter update.
As models scale, these full-size momentum states incur substantial memory overhead, motivating growing interest in memory-efficient variants of Muon.

Existing memory-efficient Muon methods have explored several routes, but remain limited in either theoretical guarantees or practical efficiency.
Subspace-based approaches establish convergence guarantees either through probabilistic full-rank correction, which can incur high peak memory~\citep{25GUM}, or under reversible-layer assumptions that do not cover standard Transformer layers~\citep{2025_SUMO}.
Momentum-compression methods reduce storage but still perform matrix-sign normalization on a full-dimensional momentum~\citep{2026_LoRAPre,2026_LiMuon}, while factorized methods require per-step factor updates~\citep{2025_TMLR_MoFaSGD}.
This raises a natural question: \emph{how can we design a memory-efficient Muon method that offers both practical efficiency and stochastic convergence guarantees for general matrix objectives?}

We pursue this goal through compact subspace optimization, which maintains momentum in a low-dimensional subspace via gradient projection.
The choice of subspace is therefore critical, as it determines the gradient information retained for parameter updates.
Under a first-order descent criterion, the ideal choice is the full-gradient TopSVD subspace, spanned by its leading singular directions.
Since the full gradient is unavailable in stochastic optimization, a natural surrogate is stochastic-gradient TopSVD, which is used by GaLore~\citep{2024_Galore}.
However, optimization over such a subspace can fail to converge under anisotropic gradient noise~\citep{25ICML_Golore,25GUM}.
GoLore~\citep{25ICML_Golore} addresses this issue by replacing TopSVD with random subspaces, whose uniform full-space coverage in expectation enables it to establish stochastic convergence guarantees for Momentum SGD (MSGD). However, it may slow practical optimization since it does not utilize dominant gradient directions.
To retain strong practical optimization while ensuring sufficient coverage, we propose a \emph{principal-random subspace} that combines leading singular directions of the stochastic gradient with random directions sampled from their orthogonal complement.

\begin{figure}[t]
    \centering
    \begin{subfigure}[t]{0.32\linewidth}
        \centering
        \begin{overpic}[width=\linewidth]{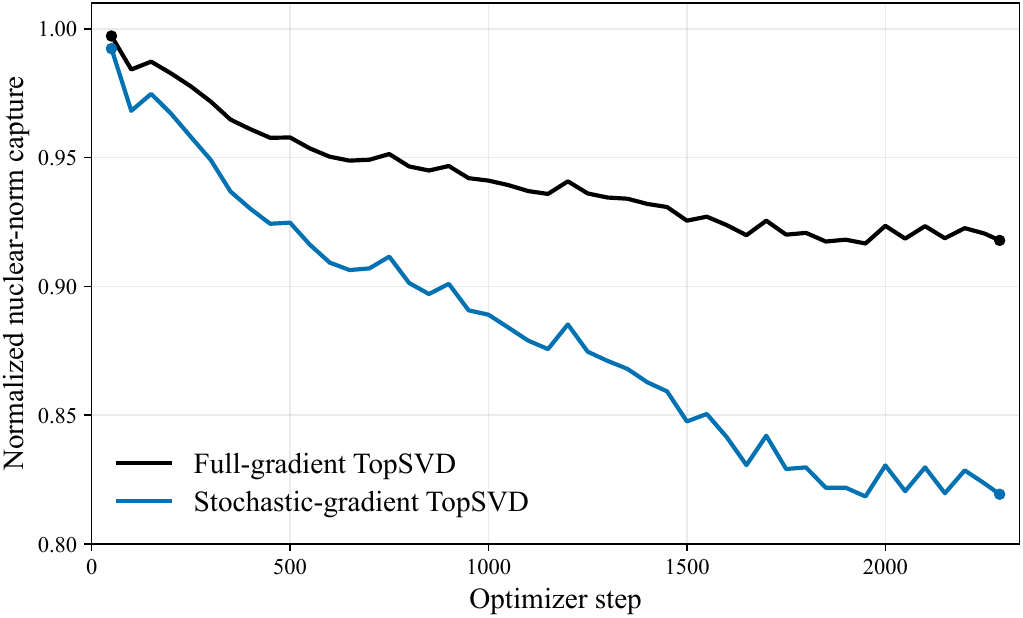}
            \put(60,50){\textbf{(a)}}
        \end{overpic}
        \label{fig:muon-nuclear-capture}
    \end{subfigure}
    \hfill
    \begin{subfigure}[t]{0.32\linewidth}
        \centering
        \begin{overpic}[width=\linewidth]{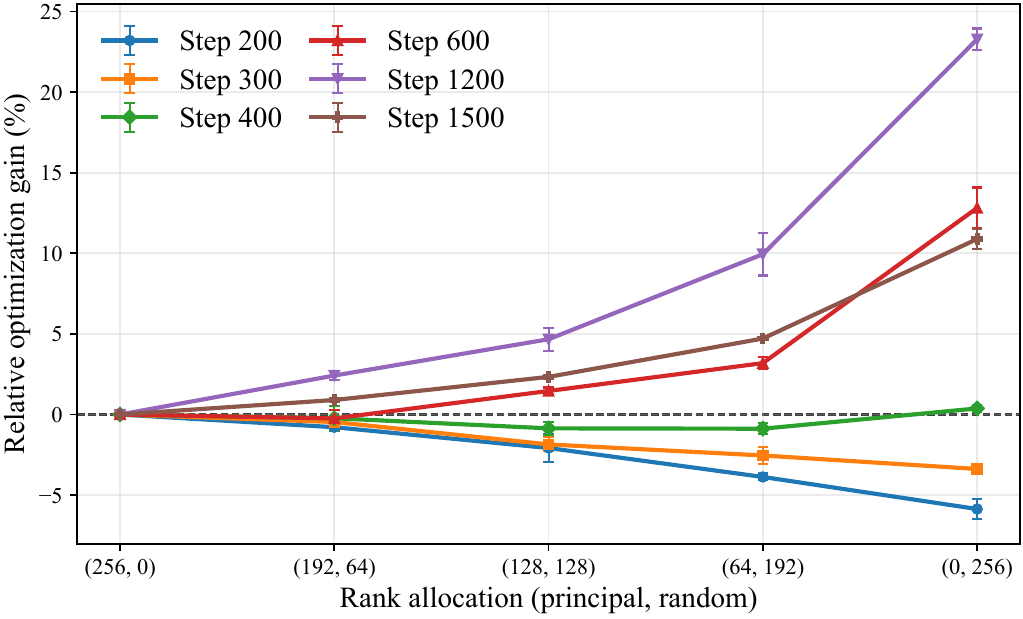}
            \put(60,50){\textbf{(b)}}
        \end{overpic}
        \label{fig:full-muon-rank-preference}
    \end{subfigure}
    \hfill
    \begin{subfigure}[t]{0.32\linewidth}
        \centering
        \begin{overpic}[width=\linewidth]{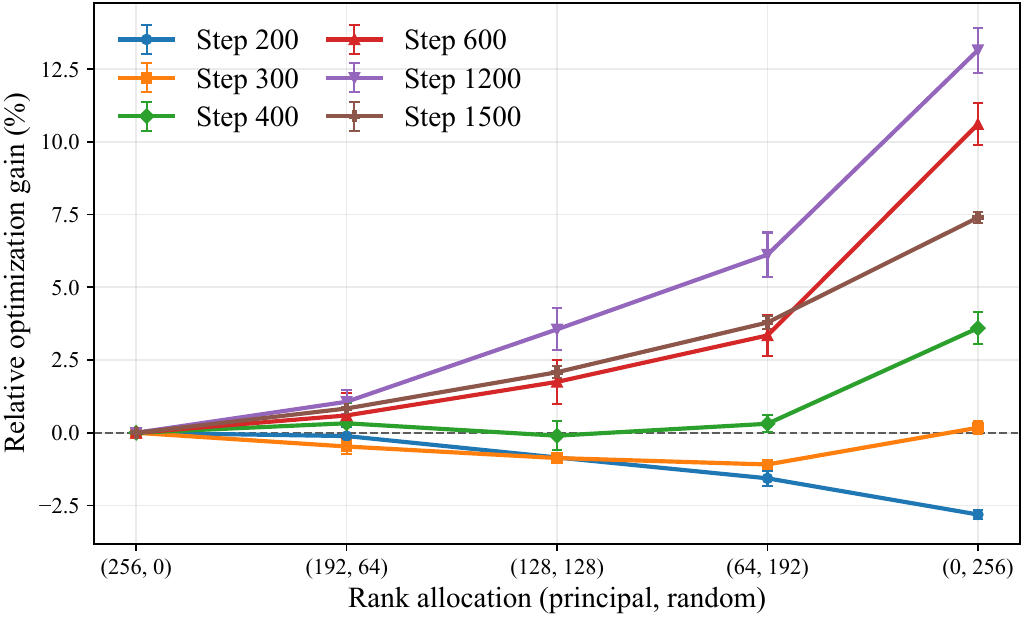}
            \put(60,50){\textbf{(c)}}
        \end{overpic}
        \label{fig:galore-muon-rank-preference}
    \end{subfigure}
    \vspace{-1em}
    \caption{LLaMA-60M pre-training. (a) Rank-256 nuclear-norm capture by the full-gradient TopSVD and stochastic-gradient TopSVD along Full-Muon. (b)--(c) Relative gain in validation-loss reduction over the principal-only allocation along Full-Muon and GaLore-Muon trajectories.}
    \label{fig:subspace-rank-transition}
    \vspace{-1em}
\end{figure}

The random directions uniformly cover the orthogonal complement of the principal subspace in expectation, which is a key property for our stochastic convergence analysis.
A remaining question is how to allocate a given rank budget to achieve strong optimization performance in practice, since stronger complementary coverage necessarily leaves fewer dimensions for principal exploitation.
Our key observation is that the preferred balance shifts over training.
Fig.~\ref{fig:subspace-rank-transition}(a) shows that even full-gradient TopSVD captures a decreasing fraction of the full-space first-order descent signal, with stochastic-gradient TopSVD incurring an additional estimation gap.
Figs.~\ref{fig:subspace-rank-transition}(b)--(c) further show that random directions are less favorable early but become beneficial later along both full-state Muon (\emph{Full-Muon}) and the GaLore-style Muon (\emph{GaLore-Muon}) trajectories.
This shift motivates reallocating the rank budget from principal toward random directions over training.

Building on the principal-random subspace, we propose \dram (\textbf{D}ynamic \textbf{R}ank \textbf{A}llocation for \textbf{M}uon), a memory-efficient Muon optimizer that reallocates rank between principal and random directions across stages while maintaining momentum and performing matrix-sign normalization entirely in the compact subspace.
We then establish convergence guarantees under general rank-allocation schedules with nondecreasing coverage factors, obtaining rates of $O(T^{-1/4})$ in the stochastic setting and $O(T^{-1/2})$ in the deterministic setting over $T$ updates. Table~\ref{tab:memory-comparison} compares \dram with related Muon methods.
With a linear reallocation schedule, \dram consistently achieves strong performance relative to low-memory baselines across both pre-training and fine-tuning settings. Our contributions are summarized as follows:
\begin{compactitem}
    \item We introduce a \emph{principal-random subspace} that preserves leading singular directions of the stochastic gradient while sampling random directions from their orthogonal complement, providing the coverage property for establishing stochastic convergence guarantees.
    \item We identify a stage-dependent shift in the preferred rank allocation for Muon from principal toward random directions over training, and propose \dram, which supports reallocating the rank budget across optimization stages.
    \item We establish convergence guarantees for \dram under general rank-allocation schedules with nondecreasing coverage factors, obtaining $O(T^{-1/4})$ and $O(T^{-1/2})$ rates in the stochastic and deterministic settings, respectively, with fixed allocation as a special case.
    \item Experiments show that \dram achieves the highest average downstream accuracy among low-memory methods at three pre-training scales and the best result across nine fine-tuning settings, with ablations supporting the proposed mechanisms.
\end{compactitem}

\section{Related Work}
We first review low-rank and subspace methods that reduce memory usage in LLM training, and then discuss memory-efficient Muon methods.

\subsection{Low-Rank and Subspace Methods}
Parameter-efficient methods achieve memory efficiency by reducing the number of trainable parameters.
LoRA~\citep{2022_lora} is a representative approach based on low-rank adapters, followed by extensions such as periodic adapter merging~\citep{2024_relora} and magnitude-direction decomposition~\citep{2024_DoRA}.
We refer readers to~\citet{2025_Lora_survey} for a broader survey.

Full-parameter subspace methods reduce optimizer-state memory by projecting gradients into low-dimensional subspaces and maintaining states therein.
GaLore~\citep{2024_Galore} constructs subspaces from the leading singular directions of stochastic gradients, whereas FLORA~\citep{2024_Flora} and GoLore~\citep{25ICML_Golore} use randomized projections.
Recent methods reduce projection overhead through sparse projections~\citep{2024_grass_NLP}, improve coverage through importance-weighted sampling~\citep{2025_SARA} or complementary direction selection~\citep{2026_ICLR_Alice}, and adapt subspaces through gradient-rank reduction~\citep{2025_ICLR_AdaRankGrad}, online PCA~\citep{24_LLM_OSD}, Grassmannian tracking~\citep{2025_NIPS_SubTrack}, and block power iteration~\citep{25_LLM_LDAdam}.
Other approaches exploit information outside projected states through norm-scaled residual updates~\citep{25_fira_NIPS}, low-rank estimation of gradient scaling~\citep{2025_APOLLO}, or hybrid optimizers~\citep{2026_ICLR_COSMOS}.
Beyond gradient-projection methods, RSO~\citep{2025_ICML_RSO} formulates training as a sequence of low-dimensional subproblems over random subspaces.

Two Adam-based methods explore direction-selection mechanisms related to our design. Alice~\citep{2026_ICLR_Alice} periodically replaces part of a tracked eigenspace with directions sampled outside the previous subspace, while keeping the rank allocation fixed.
AdaRankGrad~\citep{2025_ICLR_AdaRankGrad} adapts the rank of the gradient TopSVD subspace to exploit the decreasing effective gradient rank for memory savings, based on an information-retention threshold.
Moreover, PRAC~\citep{2026_ICML_PRAC} reduces activation memory through principal and random projections of intermediate activations. In contrast, our work focuses on reducing Muon momentum memory by performing momentum updates and matrix-sign normalization in a compact subspace.
\begin{table}[t]
\centering
\caption{Comparison of Muon methods for
$X\in\mathbb{R}^{m\times n}$ with $m\le n$. GaLore-Muon and GoLore-Muon use stochastic-gradient TopSVD and random subspaces, respectively.
$k$ denotes the subspace rank and $p$ the full-rank layer-update probability in GUM.
\dram is shown with fixed total rank.}
\vspace{-0.6em}
\label{tab:memory-comparison}
\small
\setlength{\tabcolsep}{3pt}
\begin{tabular}{cccc}
\toprule
Method & Expected storage & Worst-case storage & Guarantee \\
\midrule
Full-Muon
& $O(mn)$
& $O(mn)$
& \checkmark \\
GaLore-Muon / GoLore-Muon
& $O((m+n)k)$
& $O((m+n)k)$
& -- \\
GUM
& $O((1-p)(m+n)k+p(mk+mn))$
& $O(mn)$
& \checkmark \\
\dram (Ours)
& $O((m+n)k)$
& $O((m+n)k)$
& \checkmark \\
\bottomrule
\end{tabular}
\vspace{-0.6em}
\end{table}

\subsection{Memory-Efficient Muon}
Muon~\citep{2024_muon_jordan} has demonstrated strong performance in LLM training~\citep{2025_muon_scalable_llmtraining,2025_practical_muon_pretraining}, with recent work analyzing its convergence properties~\citep{2025_convergence_muon,2026_NS_Convergence}.
Memory-efficient Muon variants reduce optimizer-state storage through low-dimensional or low-precision representations.
GUM~\citep{25GUM} retains GaLore-style projection and establishes stochastic convergence through probabilistic full-rank correction.
However, layers selected for full-rank updates require full-size momentum states, resulting in the same worst-case storage as Full-Muon.
SUMO~\citep{2025_SUMO} performs moment orthogonalization within a dynamically adapted subspace, while its convergence guarantee is restricted to reversible layers and does not cover standard Transformer layers.
MoFaSGD~\citep{2025_TMLR_MoFaSGD} maintains a dynamically updated low-rank momentum factorization with per-step factor updates. LoRA-Pre~\citep{2026_LoRAPre} and LiMuon~\citep{2026_LiMuon} compress persistent representations, but apply matrix-sign normalization to a full-dimensional momentum.
Alternatively, 4-bit-Muon-GRASP~\citep{2026_ICLR_GRASP} and MuonQ~\citep{2026_MuonQ} reduce storage through quantization, but introduce per-step decomposition or reconstruction.
\dram differs from these methods by maintaining momentum and performing matrix-sign normalization entirely in a low-dimensional subspace while establishing stochastic convergence for general nonconvex matrix objectives.

In concurrent work, \citet{2026_PMRC} develop SMRC and PMRC, which combine subspace momentum with an orthogonal residual-gradient correction for momentum SGD and Muon, respectively, and establish stochastic convergence guarantees.
Their residual-correction approach is complementary to our principal-random subspace construction and rank reallocation.

\section{Method} \label{sec:method}
In this section, we develop our method and establish convergence guarantees. For clarity, the presentation focuses on a single matrix, and the same construction is applied to each matrix parameter.

\subsection{Compact Muon with Principal-Random Subspaces} \label{sec:principal-residual-subspace}
We consider the stochastic matrix optimization problem $\min_{X\in\mathbb{R}^{m\times n}}F(X):=\mathbb{E}_{\xi\sim\mathcal{D}}[f(X;\xi)]$, where $\mathcal{D}$ denotes the data distribution and $f(X;\xi)$ is the stochastic loss. We assume $m\le n$ without loss of generality.
When $m>n$, we apply the same construction to the transposed gradient.
Given a stochastic gradient $G_t=\nabla f(X_t;\xi_t)$, Full-Muon performs updates according to
\begin{equation}\label{eqn:update:muon}
M_t=(1-\beta)M_{t-1}+\beta G_t, \qquad X_{t+1}=X_t-\eta\,\msign(M_t),
\end{equation}
where $\beta\in(0,1]$ is the momentum coefficient, $\eta>0$ is the stepsize, and $\msign(A)=UV^\top$ for a compact SVD $A=U\Sigma V^\top$.
The full-size momentum $M_t\in\mathbb{R}^{m\times n}$ requires $O(mn)$ storage.
To reduce this cost, we divide the $T$ optimization steps into $S=\lceil T/\tau\rceil$ blocks of at most $\tau$ steps and maintain the momentum in a low-dimensional subspace that is fixed within each block. For block $s$, let $B_s\in\mathbb{R}^{m\times k_s}$ be an orthonormal basis with $1\le k_s\le m$. For $t>s\tau$ within block $s$, we update
\begin{equation} \label{eqn:update:subspace_muon}
    H_t=(1-\beta)H_{t-1}+\beta B_s^\top G_t,
    \qquad
    X_{t+1}=X_t-\eta B_s\,\msign(H_t),
\end{equation}
where $H_t\in\mathbb{R}^{k_s\times n}$ is the compact momentum.
Compared with the full-space update in~(\ref{eqn:update:muon}), (\ref{eqn:update:subspace_muon}) performs both momentum accumulation and normalization in the subspace, reducing optimizer-state storage and normalization cost.
As shown in~(\ref{eqn:update:subspace_muon}), the full-space gradient enters optimization only through its projection onto $\operatorname{Range}(B_s)$, making the construction of $B_s$ a central design choice.

To derive a criterion for subspace selection, consider the momentum-free idealized Muon update $X_{t+1}=X_t-\eta B\,\msign(B^\top\nabla F(X_t))$. The objective decrease $\Delta_t=F(X_t)-F(X_{t+1})$ satisfies
\begin{equation} \label{eqn:objective_change}
\Delta_t
=\eta\left\langle \nabla F(X_t),B\,\msign(B^\top \nabla F(X_t))\right\rangle_F+O(\eta^2)
=\eta\|B^\top \nabla F(X_t)\|_*+O(\eta^2).
\end{equation}
Thus, maximizing the first-order descent amounts to solving $\max_{B^\top B=I_k} \|B^\top\nabla F(X_t)\|_*$.
By the Ky Fan \(k\)-norm variational characterization~\citep{horn2013matrix}, the ideal rank-$k$ subspace is spanned by the leading \(k\) singular vectors of the full gradient $\nabla F(X_t)$, i.e., its TopSVD subspace.

In stochastic optimization, the full gradient is unavailable, and a natural surrogate is the TopSVD subspace of the stochastic gradient $G_t=\nabla f(X_t;\xi_t)$, which is adopted by GaLore~\citep{2024_Galore}.
However, under anisotropic gradient noise, such TopSVD-based subspace optimization can fail to converge with MSGD~\citep{25ICML_Golore} or Muon~\citep{25GUM}, as also illustrated in Section~\ref{sec:synthetic_experiment}.
The underlying issue is that the greedy TopSVD selection can become dominated by gradient noise and fail to preserve true-gradient information.
Motivated by GoLore~\citep{25ICML_Golore}, which uses random subspaces to provide full-space coverage, we retain the TopSVD directions and augment them with random directions sampled from their \emph{orthogonal complement}.

\textbf{Principal-random subspace.}
To exploit dominant stochastic-gradient components, at the beginning of block $s$, we compute $G_{s\tau}=U_{s\tau}\Sigma_{s\tau}V_{s\tau}^{\top}$ and retain the leading $r_s$ left singular directions
$R_s:=U_{s\tau}[:,1:r_s]\in\mathbb{R}^{m\times r_s}$, termed \emph{principal directions}.
To provide coverage beyond the principal subspace, we sample $q_s$ \emph{random directions} uniformly from its orthogonal complement as
\begin{equation}\label{eqn:com_projection_base}
P_s\sim \operatorname{Unif}\{P\in\mathbb{R}^{m\times q_s}:P^\top P=I_{q_s},\,R_s^\top P=0\}.
\end{equation}
In practice, $P_s$ is obtained by projecting an i.i.d.\ Gaussian matrix $Z_s\in\mathbb{R}^{m\times q_s}$ onto $\operatorname{Range}(I-R_sR_s^\top)$, followed by a sign-corrected thin QR factorization~\citep{1980_Gauss_QR,2007_correct_QR}.
The resulting basis $B_s=[R_s\;P_s]\in\mathbb{R}^{m\times(r_s+q_s)}$ spans the \emph{principal-random subspace}, reducing to the GaLore-style principal subspace~\citep{2024_Galore} when $q_s=0$ and the GoLore-style random subspace~\citep{25ICML_Golore} when $r_s=0$.
We refer to the corresponding Muon baselines as \emph{GaLore-Muon} and \emph{GoLore-Muon}, respectively, and present their update rules in Appendix~\ref{app:Implementation_Details}.
The following lemma characterizes the coverage provided by the principal-random subspace.
\begin{lemma} \label{lem:principal-preserving-coverage}
Let $R\in\mathbb{R}^{m\times r}$ satisfy $R^\top R=I_r$, and define $\Pi=RR^\top$ and $Q=I-\Pi$.
Let $P\in\mathbb{R}^{m\times q}$ be sampled as in~(\ref{eqn:com_projection_base}) with $0<q\le m-r$, and let $B=[R\;P]$ and $\delta=q/(m-r)$.
Then $\mathbb{E}[BB^\top\mid R]=\Pi+\delta Q$ and  $\mathbb{E}\!\left[\|B^\top A\|_*\mid R\right] \ge \|(\Pi+\delta Q)A\|_*\ge \delta\|A\|_*$ for any fixed $A\in\mathbb{R}^{m\times n}$.
\end{lemma}

Lemma~\ref{lem:principal-preserving-coverage} shows that the principal subspace is fully retained, while its orthogonal complement is uniformly covered in expectation with strength $\delta$.
Importantly, at block starts, the nuclear-norm bound links the projected nuclear norm $\|B^\top\nabla F(X_t)\|_*$ in the first-order descent of subspace Muon to the full-space stationarity measure $\|\nabla F(X_t)\|_*$, controlling the latter in expectation up to a factor $1/\delta$.
This connection is key to establishing the stochastic convergence guarantees in Section~\ref{sec:convergence_analysis}.

\textbf{Subspace construction cost.}
For $G\in\mathbb{R}^{m\times n}$ with $m\le n$, approximating its leading $r_s$ left singular vectors via randomized SVD costs $O(mnr_s)$~\citep{2011_Randomness_SVD}.
Given $R_s$, generating $P_s$ requires $O(mr_sq_s)$ operations for complementary projection and $O(mq_s^2)$ for thin QR factorization.
Thus, constructing $B_s$ costs $O(mnr_s+mq_sk_s)$, where $k_s=r_s+q_s$.
For a fixed total rank $k_s$, this cost interpolates between $O(mk_s^2)$ for GoLore-Muon and $O(mnk_s)$ for GaLore-Muon.
Moreover, the construction cost is amortized across the $\tau$ steps in each block.

\subsection{Dynamic Rank Allocation} \label{sec:rank-reallocation}

Under a fixed total rank, allocating more rank to random directions strengthens coverage but leaves less for principal directions, creating a trade-off between principal exploitation and complementary coverage. We next examine how this balance shifts over training and design the allocation strategy.

Recall from~(\ref{eqn:objective_change}) that $\|B^\top\nabla F(X_t)\|_*$ characterizes the first-order descent captured by the subspace spanned by $B$.
We first compare rank-256 stochastic-gradient and full-gradient TopSVD by their nuclear-norm capture relative to \(\|\nabla F(X_t)\|_*\).
Fig.~\ref{fig:subspace-rank-transition}(a) shows decreasing normalized nuclear-norm capture even for full-gradient TopSVD, with stochastic-gradient TopSVD exhibiting a widening estimation gap.
Following GoLore's observation that late-stage switching from TopSVD to random projections can improve AdamW and MSGD~\citep{25ICML_Golore}, we examine whether random directions benefit Muon over training.
At representative checkpoints, we continue training for 200 steps with fixed total rank $256$ while varying the rank allocation.
Figs.~\ref{fig:subspace-rank-transition}(b)--(c) reveal a clear stage-dependent shift, with random directions being unfavorable early but beneficial later along both Full-Muon and GaLore-Muon trajectories.
More details are provided in Section~\ref{sec:dynamic-rank-allocation}.
These results suggest that the preferred allocation should vary across training stages.

Motivated by the preference shift, we propose \dram (\textbf{D}ynamic \textbf{R}ank \textbf{A}llocation for \textbf{M}uon), which dynamically reallocates the rank budget between principal and random directions across stages.
We next specify the schedule and the corresponding compact Muon update.

\textbf{Rank allocation.}
\dram supports a general blockwise rank-allocation schedule $\{(r_s,q_s)\}_{s=0}^{S-1}$.
Consistent with the observed shift toward random directions, we focus on schedules with nondecreasing coverage factor $\delta_s=q_s/(m-r_s)$, while allowing $k_s=r_s+q_s$ to vary across blocks. This includes both gradual reallocation and hard switching, which we empirically compare in Section~\ref{sec:rank-schedule-ablation}.
Unless otherwise specified, we fix $k_s\equiv k$ and use the linear schedule $q_s=\max\{\lfloor sk/(S-1)\rfloor,1\}$ with $r_s=k-q_s$, progressively reallocating $(r_s,q_s)$ from $(k-1,1)$ to $(0,k)$ over training.

\begin{algorithm}[t]
\caption{Single-Matrix \dram}
\label{alg:DRAM}
\begin{algorithmic}[1]
\REQUIRE $X_0\in\mathbb{R}^{m\times n}$ with $m\le n$, data distribution $\D$, target horizon $T$, stepsize $\eta$, momentum parameter $\beta\in(0,1]$, block length $\tau$, and schedule $\{(r_s,q_s)\}_{s=0}^{S-1}$ with $S=\lceil T/\tau\rceil$
\FOR{$s=0,1,\ldots,S-1$}
    \FOR{$t=s\tau,s\tau+1,\ldots,\min\{(s+1)\tau,T\}-1$}
        \STATE Sample $\xi_t\sim\mathcal D$ and compute $G_t=\nabla f(X_t;\xi_t)$
        \IF{$t=s\tau$}
            \STATE Compute $G_t=U_t\Sigma_tV_t^\top$ and set $R_s=U_t[:,1:r_s]$
            \STATE Sample $P_s$ according to~(\ref{eqn:com_projection_base}) and set $B_s=[R_s\ P_s]$
        \ENDIF
        \STATE Compute $\widehat G_t=B_s^\top G_t$ and update $H_t$ by~(\ref{eqn:DRAM_Muon_Ht})
        \STATE Update $X_{t+1}=X_t-\eta B_s\,\msign(H_t)$
    \ENDFOR
\ENDFOR
\RETURN $X_T$.
\end{algorithmic}
\end{algorithm}

\textbf{\dram update.}
At the beginning of block $s$, \dram constructs $B_s\in\mathbb{R}^{m\times k_s}$ based on the scheduled allocation $(r_s,q_s)$ as in Section~\ref{sec:principal-residual-subspace}.
For each stochastic gradient $G_t=\nabla f(X_t;\xi_t)$, \dram projects it into the subspace as $\widehat G_t=B_s^\top G_t\in\mathbb{R}^{k_s\times n}$, and updates the compact momentum by
\begin{equation}
\label{eqn:DRAM_Muon_Ht}
H_t=
\begin{cases}
\beta\widehat G_t, & t=s\tau,\\
(1-\beta)H_{t-1}+\beta\widehat G_t, & t>s\tau.
\end{cases}
\end{equation}
\dram then applies the matrix sign to $H_t$ in the compact subspace and lifts it back to the full space to update $X_{t+1}=X_t-\eta B_s\,\msign(H_t).$
The complete procedure is summarized in Algorithm~\ref{alg:DRAM}.

\textbf{Memory and normalization cost.}
In each block $s$, \dram stores the projection basis $B_s\in\mathbb{R}^{m\times k_s}$ and compact momentum $H_t\in\mathbb{R}^{k_s\times n}$, requiring $O((m+n)k_s)$ persistent storage. With the same rank, GaLore-Muon and GoLore-Muon have the same order of storage cost.
GUM~\citep{25GUM} also maintains compact states for low-rank updates, but a layer sampled for full-rank correction stores an $m\times k$ projection matrix and an $m\times n$ momentum, yielding $O(mn)$ worst-case storage.
Moreover, \dram computes the matrix-sign normalization directly on $H_t$, reducing the cost of each Newton--Schulz iteration from $O(m^2n)$ in the full space to $O(k_s^2n)$ in the compact subspace.

\subsection{Convergence Analysis}   \label{sec:convergence_analysis}
We analyze \dram under the operator/nuclear-norm geometry used in prior Muon analyses~\citep{25GUM,2026_NS_Convergence,2026_NeurIPS_SignHeavyTails,2026_NeurIPS_SignSGD_LowerBounds}, with the following assumptions.
\begin{assumption} \label{ass:lower_bound}
The objective function is lower bounded, i.e., $F^*:=\inf_X F(X)>-\infty$.
\end{assumption}
\begin{assumption}\label{ass:gradient_filtration}
Let $\{\mathcal F_t\}_{t\ge0}$ be the natural filtration generated by all randomness before sampling $\xi_t$. For every $t$, $\E[\nabla f(X_t;\xi_t)\mid\mathcal F_t]=\nabla F(X_t)$ and $\E[\|\nabla f(X_t;\xi_t)-\nabla F(X_t)\|_F^2\mid\mathcal F_t]\le\sigma^2$.
\end{assumption}
\begin{assumption} \label{ass:nuclear_smoothness}
The objective $F:\mathbb{R}^{m\times n}\to\mathbb{R}$ is $L_*$-smooth with respect to the operator/nuclear norm pair, namely, for any $X,Y\in\mathbb{R}^{m\times n}$, $\|\nabla F(X)-\nabla F(Y)\|_*\le L_*\|X-Y\|_{\op}$.
\end{assumption}
\begin{assumption}\label{ass:exact_muon}
For the theoretical analysis, we represent Muon normalization by the exact matrix sign map $\,\msign(A)=UV^\top$ for an SVD $A=U\Sigma V^\top$.
\end{assumption}
Under stochastic gradients, the complementary coverage factor $\delta_s=q_s/(m-r_s)$ from Lemma~\ref{lem:principal-preserving-coverage} relates subspace descent to full-gradient stationarity.
In the deterministic setting, we obtain the full gradient and the coverage factor becomes $\gamma_s=k_s/m$ with $k_s=r_s+q_s$. In this case, random directions are unnecessary and $q_s$ can be set to zero.
We establish convergence guarantees for stage-dependent rank-allocation schedules with nondecreasing coverage factors as follows.

\begin{theorem}\label{thm:dram_convergence}
Suppose Assumptions~\ref{ass:lower_bound}--\ref{ass:exact_muon} hold and let $\Delta:=F(X_0)-F^*>0$. Choose parameters specified in
Appendix~\ref{app:dynamic_convergence}, and then prescribe a schedule $\{(r_s,q_s)\}_{s=0}^{S-1}$ independently of the optimization randomness. For $\sigma>0$, fix a rank cap $1\le k_{\max}\le m$, and assume $0\le r_s<m$, $1\le q_s\le m-r_s$, $k_s=r_s+q_s\le k_{\max}$, and $\{\delta_s\}_{s=0}^{S-1}$ is nondecreasing. For $\sigma=0$, assume $0\le r_s\le m$, $0\le q_s\le m-r_s$, $k_s>0$, and $\{\gamma_s\}_{s=0}^{S-1}$ is nondecreasing. Then, Algorithm~\ref{alg:DRAM} satisfies
\[
\frac{1}{T}\sum_{t=0}^{T-1}\E\|\nabla F(X_t)\|_*=
\begin{cases}O\!\left(\Gamma_\delta\left[\sqrt{\frac{L_*\Delta}{T}}+\left(\frac{k_{\max}L_*\Delta\sigma^2}{T}\right)^{1/4}\right]\right),&\sigma>0,\\[1mm]
O\!\left(\Gamma_\gamma\sqrt{\frac{L_*\Delta}{T}}\right),&\sigma=0.\end{cases}
\]
where $\Gamma_\delta:=\delta_0^{-1}+\frac{1}{S}\sum_{s=0}^{S-1}\delta_s^{-1}$ and $\Gamma_\gamma:=\gamma_0^{-1}+\frac{1}{S}\sum_{s=0}^{S-1}\gamma_s^{-1}$.
\end{theorem}
\begin{corollary}\label{cor:dram_fixed_allocation}
Suppose the conditions of Theorem~\ref{thm:dram_convergence} hold with $r_s\equiv r$ and $q_s\equiv q$, and let $k=r+q$. For $\sigma>0$, $\frac1T\sum_{t=0}^{T-1}\E\|\nabla F(X_t)\|_*=O(\delta^{-1}[\sqrt{L_*\Delta/T} +(L_*\Delta k\sigma^2/T)^{1/4}])$ with $\delta=q/(m-r)$. For $\sigma=0$ and $k>0$, $\frac1T\sum_{t=0}^{T-1}\E\|\nabla F(X_t)\|_*=O(m\sqrt{L_*\Delta}/(k\sqrt{T}))$.
\end{corollary}
\textbf{Remark.}
Theorem~\ref{thm:dram_convergence} establishes $O(T^{-1/4})$ stochastic and $O(T^{-1/2})$ deterministic convergence rates up to rank-schedule factors, matching the corresponding orders for Muon~\citep{2026_NS_Convergence}.

\section{Experiments} \label{sec:experiment}
We evaluate \dram through synthetic optimization, stage-wise rank-allocation analysis, LLM pre-training, full-parameter fine-tuning, and rank-allocation schedule ablations.

\subsection{Synthetic Settings}\label{sec:synthetic_experiment}

Following prior work~\citep{25ICML_Golore,25GUM}, we consider a quadratic problem with anisotropic gradient noise:
\begin{equation}
    \min_{X\in\R^{n\times n}}
    F(X)=\frac{1}{2}\|AX\|_F^2 + \langle B, X\rangle_F,
    \qquad
    \nabla f(X;\xi)=A^\top AX+B+\xi\sigma C,
\end{equation}
where $A=[I_{n-h},0]\in\R^{(n-h)\times n}$, $B=\begin{bmatrix}D & 0\\0 & 0\end{bmatrix}\in\R^{n\times n}$ and $C=\begin{bmatrix}0 & 0\\0 & I_h\end{bmatrix}\in\R^{n\times n}$.
Here \(D\in\R^{(n-h)\times(n-h)}\) has i.i.d. standard Gaussian entries, $\xi\sim\operatorname{Unif}\{-1,1\}$, and $\sigma$ controls the noise level. We set $n=20$, $h=12$, and $\sigma=1$. The true gradient and noise lie in orthogonal subspaces.

\begin{wrapfigure}{r}{0.4\linewidth}
\vspace{-1\baselineskip}
\centering
\captionsetup{skip=4.5pt}
\includegraphics[width=\linewidth]{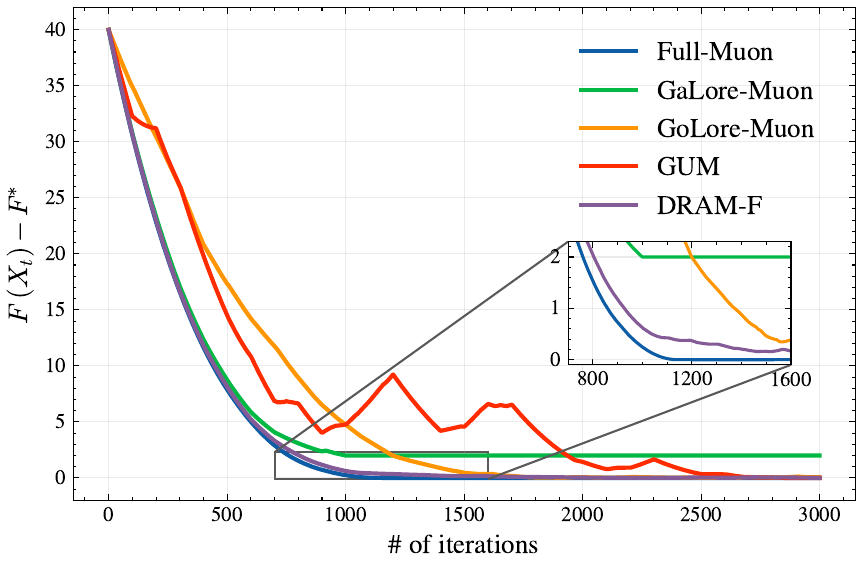}
\caption{Synthetic optimization.}
\label{fig:synthetic_loss}
\vspace{-1.2\baselineskip}
\end{wrapfigure}
\textbf{Algorithms.}
We compare the fixed-allocation variant \dramf with Full-Muon, GaLore-Muon, GoLore-Muon, and GUM. All methods run for 3000 iterations with a learning rate of \(5\times10^{-3}\). Subspace methods refresh their projections every 100 iterations. GaLore-Muon and GoLore-Muon use rank $8$, and \dramf uses $(r,q)=(6,2)$. GUM uses rank $4$ and full-rank correction probability $p=0.5$ to match the expected storage cost.

\textbf{Results.}
Fig.~\ref{fig:synthetic_loss} plots $F(X_t)-F^*$, where $F^*=-\frac{1}{2}\|D\|_F^2$ is the optimum. Full-Muon converges rapidly and serves as a reference trajectory, while the low-memory baselines exhibit different limitations.
GaLore-Muon decreases the error quickly at first but plateaus at a nonzero gap. GoLore-Muon converges more slowly, whereas GUM shows a more oscillatory trajectory.
In contrast, \dramf converges stably at a speed comparable to Full-Muon and outperforms the other low-memory methods, demonstrating the benefit of combining principal directions for fast optimization with complementary coverage to maintain descent under anisotropic noise.

\subsection{Stage-Wise Rank Allocation Analysis} \label{sec:dynamic-rank-allocation}

Along a Full-Muon LLaMA-60M pre-training trajectory on C4~\citep{2020_C4_dataset}, we measure normalized nuclear-norm capture $C_t(B)=\sum_{\ell}\|B_{t,\ell}^{\top}\bar G_{t,\ell}\|_*/\sum_{\ell}\|\bar G_{t,\ell}\|_*$, where $\ell$ indexes all matrix parameters, with full gradients $\bar G_{t,\ell}$ and projection bases $B_{t,\ell}$.
Under the idealized Muon update in~(\ref{eqn:objective_change}), $C_t(B)$ measures the fraction of full-space first-order descent retained by subspace optimization.
We compare rank-$256$ full-gradient and stochastic-gradient TopSVD, averaging the latter over 64 independent mini-batches.
Fig.~\ref{fig:subspace-rank-transition}(a) shows that their capture drops from $0.997$ to $0.918$ and from $0.992$ to $0.819$, respectively, revealing reduced fixed-rank capture and an increased stochastic estimation gap.
To test rank-allocation preferences, we run $200$-step continuations from Full-Muon and GaLore-Muon checkpoints with $r+q=256$ and $q\in\{0,64,128,192,256\}$.
Relative gain over the principal-only baseline is $\mathcal{G}_{t}(r,q)=(L_{t}^{(256,0)}(200)-L_{t}^{(r,q)}(200))/(L_t(0)-L_{t}^{(256,0)}(200))\times100\%$, where $L_t(0)$ and $L_t^{(r,q)}(200)$ are validation losses before and after continuation.
Using this metric, Figs.~\ref{fig:subspace-rank-transition}(b)--(c) show that allocating more rank to random directions is unfavorable early but becomes beneficial later, with the mean $\pm$ one standard deviation reported over three seeds. This stage-dependent shift supports progressive rank reallocation from principal to random directions.
Results for total ranks $64$ and $128$ are provided in Appendix~\ref{app:stagewise-rank-details}.

\subsection{LLM Pre-training} \label{sec:llm-pretraining}
\begin{table}[t]
\centering
\small
\setlength{\tabcolsep}{3.3pt}
\caption{
Zero-shot downstream accuracy (\%) of models pre-trained on C4.
Dashed lines separate full-memory and low-memory methods. In the Avg.\ column, the best and second-best results among low-memory methods are highlighted in \textbf{bold} and \underline{underlined}, respectively.
}
\label{tab:pre_train}
\begin{tabular}{@{}llcccccccc@{}}
\toprule
Model & Method & ARC-E & ARC-C & OBQA & HellaSwag & PIQA & SIQA & WinoGrande & Avg. \\
\midrule
\multirow{7}{*}{LLaMA-60M}
& AdamW & 29.67 & 21.50 & 28.20 & 26.38 & 58.49 & 35.72 & 50.67 & 35.80 \\
& Full-Muon & 32.37 & 22.10 & 27.00 & 26.33 & 59.52 & 35.52 & 52.49 & 36.47 \\

\cdashline{2-10}
\noalign{\vskip 2pt}
& GaLore-Muon & 31.19 & 20.48 & 26.20 & 26.02 & 58.87 & 35.77 & 53.04 & 35.94 \\
& GoLore-Muon & 31.10 & 19.97 & 28.00 & 26.16 & 59.52 & 35.52 & 51.62 & \underline{35.98} \\
& GUM & 31.14 & 20.22 & 27.80 & 25.93 & 58.76 & 33.78 & 52.49 & 35.73 \\
& \dramf & 31.48 & 20.48 & 28.40 & 25.98 & 59.30 & 34.54 & 51.70 & \underline{35.98} \\
& \dram & 30.81 & 21.50 & 28.80 & 26.04 & 58.98 & 35.06 & 52.80 & \textbf{36.28} \\
\midrule
\multirow{7}{*}{LLaMA-130M}
& AdamW & 34.34 & 21.84 & 28.20 & 28.30 & 61.37 & 36.18 & 52.41 & 37.52 \\
& Full-Muon & 34.13 & 22.10 & 27.60 & 28.82 & 62.24 & 36.08 & 50.28 & 37.32 \\
\cdashline{2-10}
\noalign{\vskip 2pt}
& GaLore-Muon & 33.21 & 21.84 & 27.60 & 27.16 & 60.94 & 35.16 & 51.85 & 36.82 \\
& GoLore-Muon & 33.54 & 20.73 & 29.80 & 27.07 & 59.90 & 35.36 & 50.99 & 36.77 \\
& GUM & 33.84 & 21.84 & 28.20 & 27.36 & 60.12 & 35.88 & 52.57 & \underline{37.11} \\
& \dramf & 33.29 & 21.67 & 27.80 & 27.19 & 61.37 & 36.23 & 51.62 & 37.02 \\
& \dram & 33.12 & 22.61 & 29.00 & 26.94 & 60.39 & 34.80 & 53.43 & \textbf{37.19} \\
\midrule
\multirow{7}{*}{LLaMA-350M}
& AdamW & 36.95 & 22.70 & 27.80 & 33.83 & 66.76 & 37.67 & 53.35 & 39.87 \\
& Full-Muon & 37.92 & 23.38 & 27.40 & 34.88 & 66.92 & 38.64 & 52.33 & 40.21 \\
\cdashline{2-10}
\noalign{\vskip 2pt}
& GaLore-Muon & 34.72 & 22.61 & 27.20 & 30.49 & 63.93 & 36.18 & 50.36 & 37.93 \\
& GoLore-Muon & 36.03 & 22.61 & 26.00 & 30.33 & 63.22 & 36.64 & 52.33 & 38.17 \\
& GUM & 36.20 & 22.10 & 27.20 & 30.80 & 63.87 & 36.34 & 52.64 & \underline{38.45} \\
& \dramf & 35.44 & 22.78 & 28.20 & 30.42 & 63.76 & 35.77 & 51.22 & 38.23 \\
& \dram & 34.93 & 23.89 & 28.20 & 30.45 & 63.55 & 37.31 & 51.54 & \textbf{38.55} \\
\bottomrule
\end{tabular}
\vspace{-0.2em}
\end{table}

\textbf{Setup.}
We pre-train LLaMA-60M, 130M, and 350M from scratch on C4~\citep{2020_C4_dataset} for 1.2B, 2.5B, and 7B tokens, respectively. Following~\citet{25GUM}, we evaluate zero-shot accuracy on ARC~\citep{eval_2018ARC}, OpenBookQA~\citep{eval_2018OpenBookQA}, HellaSwag~\citep{eval_2019hellaswag}, PIQA~\citep{eval_2020PIQA}, SIQA~\citep{eval_2019socialIQa}, and WinoGrande~\citep{eval_2021WinoGrande}.
We compare \dram with AdamW and Full-Muon as full-memory references and with GaLore-Muon, GoLore-Muon, and GUM as low-memory baselines.
GaLore-Muon and GoLore-Muon use rank 256. \dramf fixes $(r,q)=(128,128)$, while \dram linearly reallocates $(r,q)$ from $(255,1)$ to $(0,256)$ over training. GUM uses rank 128, with the full-rank correction probability chosen to match the expected storage of the other low-memory methods.
All Muon-based methods are tuned over the same scale-specific learning-rate grids. More details are provided in Appendix~\ref{app:pretraining-hyperparameters}.

\textbf{Results.}
Table~\ref{tab:pre_train} shows that \dram achieves the highest average downstream accuracy among low-memory methods at all three model scales.
With a fixed allocation, \dramf matches or outperforms both GaLore-Muon and GoLore-Muon in average downstream accuracy at all three scales, suggesting that combining principal and random directions can be beneficial relative to either endpoint.
Moreover, \dram consistently improves over \dramf across the three scales, supporting the benefit of stage-wise rank reallocation.
Overall, these results demonstrate the effectiveness of both the principal-random subspace design and stage-wise rank reallocation during pre-training.

\subsection{LLM Fine-tuning} \label{sec:llm-finetuning}
\textbf{Setup.}
We perform full-parameter supervised fine-tuning of Qwen3-8B~\citep{2025_Qwen3}, LLaMA-3.1-8B~\citep{2024_Llama3}, and Gemma-2-9B~\citep{2024_Gemma2} on instruction following, mathematical reasoning, and code generation.
We fine-tune each model for one epoch on GPT4-LLM-Cleaned~\citep{2023_GPT4_Offical}, GSM8K~\citep{2021_GSM8K}, and MBPP~\citep{2021_MBPP}, and evaluate on IFEval~\citep{2023_IFEval} with strict prompt-level accuracy, GSM8K with accuracy, and MBPP with Pass@1.
We compare \dram with Full-Muon, GaLore-Muon, GoLore-Muon, and GUM. \dram, GaLore-Muon, and GoLore-Muon use rank 512, while GUM uses rank 128 with a model-specific full-rank correction probability set to match the expected storage of these low-memory methods. Additional details are provided in Appendix~\ref{app:finetuning-hyperparameters}.

\begin{wraptable}{r}{0.46\linewidth}
\vspace{-1\baselineskip}
\centering
\caption{Peak GPU memory (GiB) on MBPP.}
\label{tab:memory_main}
\small
\setlength{\tabcolsep}{3pt}
\begin{tabular}{lcc}
\toprule
Backbone & Full $\rightarrow$ \dram & Reduction \\
\midrule
Qwen3-8B
& $52.80\!\rightarrow\!42.47$ & $19.6\%$ \\
LLaMA-3.1-8B
& $50.81\!\rightarrow\!40.31$ & $20.7\%$ \\
Gemma-2-9B
& $60.25\!\rightarrow\!48.06$ & $20.2\%$ \\
\bottomrule
\end{tabular}
\vspace{-0.6\baselineskip}
\end{wraptable}
\textbf{Results.}
Table~\ref{tab:memory_main} shows the practical memory savings on MBPP, where \dram reduces peak GPU memory by $19.6$--$20.7\%$ relative to Full-Muon across the three backbones.
The complete results in Appendix~\ref{app:memory} show reductions of $17.9$--$20.7\%$ across all nine settings.
Despite this memory reduction, Table~\ref{tab:fine_tune} shows that \dram achieves the best low-memory result in eight of the nine settings and ties for the best in the remaining one.
It also matches or exceeds Full-Muon in eight of nine settings.

\begin{table}[t]
\centering
\small
\setlength{\tabcolsep}{7pt}
\caption{
Fine-tuning performance (\%) across three tasks and three model backbones.
Dashed lines separate full-memory and low-memory methods. The best and second-best results among low-memory methods are highlighted in \textbf{bold} and \underline{underlined}, respectively.
}
\label{tab:fine_tune}
\vspace{-0.3em}
\begin{tabular}{llccc}
\toprule
Model & Method
& \makecell{Instruction following}
& \makecell{Math}
& \makecell{Code generation} \\
\midrule
\multirow{5}{*}{Qwen3-8B}
& Full-Muon    & 50.46 & 84.16 & 71.60 \\
\cdashline{2-5}
\noalign{\vskip 2pt}
& GaLore-Muon  & \underline{50.28} & 83.47 & 71.80 \\
& GoLore-Muon  & 47.87 & 83.47 & \underline{72.00} \\
& GUM     & 48.98 & \underline{83.62} & 71.80 \\
& \dram & \textbf{50.46} & \textbf{84.16} & \textbf{72.20} \\
\midrule
\multirow{5}{*}{LLaMA-3.1-8B}
& Full-Muon    & 31.79 & 52.77 & 49.40 \\
\cdashline{2-5}
\noalign{\vskip 2pt}
& GaLore-Muon  & \underline{30.87} & 49.89 & \underline{49.40} \\
& GoLore-Muon  & 30.68 & 48.67 & 49.20 \\
& GUM     & 30.13 & \underline{50.27} & \textbf{50.20} \\
& \dram & \textbf{31.79} & \textbf{51.02} & \textbf{50.20} \\
\midrule
\multirow{5}{*}{Gemma-2-9B}
& Full-Muon    & 38.45 & 75.06 & 52.40 \\
\cdashline{2-5}
\noalign{\vskip 2pt}
& GaLore-Muon  & 38.82 & 74.68 & 54.00 \\
& GoLore-Muon  & 39.00 & 73.31 & 54.20 \\
& GUM     & \underline{39.37} & \underline{74.83} & \underline{54.40} \\
& \dram & \textbf{40.11} & \textbf{75.21} & \textbf{55.20} \\
\bottomrule
\end{tabular}
\vspace{-0.5em}
\end{table}

\subsection{Ablation on Rank-Allocation Schedules}\label{sec:rank-schedule-ablation}
\textbf{Setup.}
We fix the total rank at 256 and compare Principal-only, Random-only, a fixed $(128,128)$ allocation, hard-switch schedules, and the default Linear schedule.
Each hard-switch schedule starts from $(255,1)$ and switches directly to $(0,256)$ at 25\%, 50\%, or 75\% of training, while Linear gradually reallocates rank between the same endpoints.
Table~\ref{tab:rank_schedule_ablation} reports the validation-loss improvement over Principal-only, averaged over three random seeds.
All allocations in Figs.~\ref{fig:subspace-rank-transition}(b)--(c) and~\ref{fig:rank-allocation-rank-robustness} and Table~\ref{tab:rank_schedule_ablation} use the same momentum-resetting rule in~(\ref{eqn:DRAM_Muon_Ht}), including Principal-only and Random-only.

\textbf{Results.}
First, Random-only yields higher validation loss than Principal-only on both model sizes, showing that coverage alone is insufficient in these settings.
Second, the fixed $(128,128)$ allocation provides little benefit on LLaMA-60M and degrades LLaMA-130M, suggesting that a static allocation is not consistently effective.
Switch-25\% has a larger mean gain than Linear at both scales, but Switch-50\% and Switch-75\% do worse. Switching at 0\% recovers Random-only, so earlier switching is not uniformly better. Linear is a simple default that improves both scales without tuning a switch point.

\begin{table}[t]
\centering
\small
\setlength{\tabcolsep}{3.5pt}
\caption{
Rank-schedule ablation on pre-training validation loss.
$\Delta L_{\rm val}=L_{\rm Principal}-L_{\rm schedule}$, where larger values indicate greater improvement over Principal-only.
We report the mean $\pm$ standard deviation over three random seeds.
Realloc.\ denotes reallocation; HP denotes hyperparameter.
}
\label{tab:rank_schedule_ablation}
\vspace{-0.3em}
\begin{tabular}{lccccc}
\toprule
& \multicolumn{3}{c}{Schedule properties}
& \multicolumn{2}{c}{$\Delta L_{\rm val}$ $(\times10^{-3})\uparrow$} \\
\cmidrule(lr){2-4}\cmidrule(lr){5-6}
Schedule
& \makecell{Coverage}
& \makecell{Rank realloc.}
& \makecell{No schedule HP}
& LLaMA-60M & LLaMA-130M \\
\midrule
Principal-only
& -- & -- & \checkmark
& $0.0$ (ref.) & $0.0$ (ref.) \\
Random-only
& \checkmark & -- & \checkmark
& $-7.7$\std{1.4} & $-8.9$\std{2.7} \\
Fixed $(128,128)$
& \checkmark & -- & \checkmark
& $+0.1$\std{1.4} & $-2.2$\std{0.4} \\
Switch-25\%
& \checkmark & \checkmark & --
& $+1.6$\std{0.4} & $+1.4$\std{0.3} \\
Switch-50\%
& \checkmark & \checkmark & --
& $+0.5$\std{0.2} & $0.0$\std{0.2} \\
Switch-75\%
& \checkmark & \checkmark & --
& $0.0$\std{0.1} & $-0.3$\std{0.03} \\
\textbf{Linear} (Default)
& \checkmark & \checkmark & \checkmark
& $+1.2$\std{1.0} & $+0.9$\std{0.05} \\
\bottomrule
\end{tabular}
\vspace{-0.5em}
\end{table}

\subsection{Memory and Training Efficiency}\label{sec:efficiency-comparison}
\textbf{Setup.}
We benchmark Qwen3-8B fine-tuning on GSM8K using Full-Muon, 8bit-Muon~\citep{2026_quantization_muon}, 4bit-Muon-GRASP~\citep{2026_ICLR_GRASP}, MuonQ~\citep{2026_MuonQ}, SUMO~\citep{2025_SUMO}, MoFaSGD~\citep{2025_TMLR_MoFaSGD}, and \dram.
Each run uses a single RTX PRO 5000 72GB GPU, batch size 1, and sequence length 1024.
Models and gradients use BF16, while SVD/QR workspaces use FP32.
The three low-rank methods (\dram, SUMO, and MoFaSGD) use rank 512 and BF16 persistent states.
Optimizer-state memory includes the shared 4.638 GiB of AdamW states for parameters not updated by Muon.
Throughput includes all training-step computation over 100 timed updates after 10 warmup updates, including two subspace refreshes for \dram ($\tau=50$).
\begin{table}[t]
\centering
\small
\setlength{\tabcolsep}{3pt}
\caption{Memory and training efficiency of Qwen3-8B fine-tuning on GSM8K. Values are means $\pm$ standard deviations over three runs; throughput counts padded tokens.}
\label{tab:rebuttal-efficiency}
\begin{tabular}{lccc}
\toprule
Method & Optimizer state (GiB)$\downarrow$ & Peak GPU memory (GiB)$\downarrow$ & \makecell{Padded throughput\\(tokens/s)$\uparrow$} \\
\midrule
Full-Muon & $17.575$\std{0.000} & $52.802$\std{0.000} & $218.765$\std{0.767} \\
8bit-Muon & $11.119$\std{0.000} & $46.346$\std{0.000} & $200.721$\std{0.038} \\
4bit-Muon-GRASP & $8.899$\std{0.000} & $44.285$\std{0.000} & $147.267$\std{0.088} \\
MuonQ & $8.165$\std{0.000} & $43.454$\std{0.000} & $146.391$\std{0.033} \\
SUMO & $7.239$\std{0.000} & $42.466$\std{0.000} & $75.997$\std{4.861} \\
MoFaSGD & $7.240$\std{0.000} & $42.466$\std{0.000} & $41.545$\std{0.193} \\
\textbf{\dram} & $7.239$\std{0.000} & $42.466$\std{0.000} & $223.170$\std{2.344} \\
\bottomrule
\end{tabular}
\end{table}

\textbf{Results.}
In this benchmark, \dram uses 34.9\% less optimizer-state memory and 8.4\% less peak GPU memory than 8bit-Muon, with 11.2\% higher throughput.
Relative to 4bit-Muon-GRASP and MuonQ, its peak-memory reductions are 4.1\% and 2.3\%, with throughput improvements of 51.5\% and 52.4\%, respectively.
At nearly identical memory usage, \dram achieves $2.94\times$ SUMO's and $5.37\times$ MoFaSGD's throughput.
DRAM combines matrix-sign normalization on only the compact $k\times n$ momentum with amortized basis construction over 50 steps, whereas SUMO uses a compact SVD each step and MoFaSGD uses two QR factorizations and a core SVD each step.

\section{Conclusion}
We introduce \dram, a memory-efficient Muon optimizer built on a principal-random subspace with stage-wise rank reallocation between principal and random directions.
This design achieves strong practical performance with provable convergence, while keeping momentum and normalization in a compact subspace.
We establish \(O(T^{-1/4})\) stochastic and \(O(T^{-1/2})\) deterministic convergence rates and demonstrate strong performance across pre-training and fine-tuning.

\bibliographystyle{iclr2027_conference}
\bibliography{ref}

\appendix
\section{Additional Experimental Details}
\subsection{Implementation Details}
\label{app:Implementation_Details}
For randomized basis generation, we use a sign-corrected QR factorization~\citep{2007_correct_QR}.
For \dram with $q_s>0$, to sample $P_s$ according to~(\ref{eqn:com_projection_base}), we draw a fresh $Z_s\in\mathbb{R}^{m\times q_s}$ with i.i.d.\ $\mathcal{N}(0,1)$ entries and project it onto the orthogonal complement of $R_s$ as $\widetilde Z_s=Z_s-R_s(R_s^\top Z_s).$ We then compute a thin QR factorization $\widetilde Z_s=\widetilde P_s T_s$ and apply the standard sign correction to obtain $P_s$. For $q_s=0$, we set $P_s$ empty and skip sampling and QR.
This produces a Haar-uniform orthonormal $q_s$-frame in $\operatorname{Range}(I-R_sR_s^\top)$.
For GoLore-Muon, we apply the same procedure to an i.i.d.\ Gaussian matrix of the target rank.
For completeness, we present the Muon instantiations of two representative subspace baselines, namely GaLore-Muon in Algorithm~\ref{alg:galore_muon} and GoLore-Muon in Algorithm~\ref{alg:golore_muon}.
At subspace refreshes, GaLore-Muon and GoLore-Muon follow the momentum-state handling conventions of their respective implementations~\citep{2024_Galore,25ICML_Golore}. GaLore-Muon retains the compact momentum when replacing the principal basis, while GoLore-Muon transports the previous compact state into the new basis through $B_s^\top B_{s-1}$.

All pre-training experiments use four GPUs with PyTorch DistributedDataParallel and NCCL, while all fine-tuning experiments use a single GPU.
The pre-training and fine-tuning quality experiments were conducted on NVIDIA RTX PRO 6000 Blackwell Server Edition GPUs, running Ubuntu 22.04 with PyTorch and CUDA 13.0.

\begin{algorithm}[t]
\caption{Single-Matrix GaLore-Muon}
\label{alg:galore_muon}
\begin{algorithmic}[1]
\REQUIRE Initial point $X_0\in\mathbb{R}^{m\times n}$ with $m\le n$, data distribution $\mathcal D$, target horizon $T$, stepsize $\eta$, momentum parameter $\beta\in(0,1]$, projection rank $k$, and block length $\tau$, with $S=\lceil T/\tau\rceil$
\STATE Initialize $H_{-1}=0$.
\FOR{$s=0,1,\ldots,S-1$}
    \FOR{$t=s\tau,s\tau+1,\ldots,\min\{(s+1)\tau,T\}-1$}
        \STATE Sample $\xi_t\sim\mathcal D$ and compute $G_t=\nabla f(X_t;\xi_t)$
        \IF{$t=s\tau$}
            \STATE Compute $G_t=U_t\Sigma_tV_t^\top$ and set $B_s=U_t[:,1:k]$
        \ENDIF
        \STATE Compute $\widehat G_t=B_s^\top G_t$
        \STATE Update $H_t=(1-\beta)H_{t-1}+\beta\widehat G_t$
        \STATE Update $X_{t+1}=X_t-\eta B_s\,\msign(H_t)$
    \ENDFOR
\ENDFOR
\RETURN $X_T$.
\end{algorithmic}
\end{algorithm}

\begin{algorithm}[t]
\caption{Single-Matrix GoLore-Muon}
\label{alg:golore_muon}
\begin{algorithmic}[1]
\REQUIRE Initial point $X_0\in\mathbb{R}^{m\times n}$ with $m\le n$, data distribution $\mathcal D$, target horizon $T$, stepsize $\eta$, momentum parameter $\beta\in(0,1]$, projection rank $k$, and block length $\tau$, with $S=\lceil T/\tau\rceil$
\STATE Initialize $H_{-1}=0$.
\FOR{$s=0,1,\ldots,S-1$}
    \FOR{$t=s\tau,s\tau+1,\ldots,\min\{(s+1)\tau,T\}-1$}
        \STATE Sample $\xi_t\sim\mathcal D$ and compute $G_t=\nabla f(X_t;\xi_t)$
        \IF{$t=s\tau$}
            \STATE Sample $B_s\sim\operatorname{Unif}\{B\in\mathbb{R}^{m\times k}:B^\top B=I_k\}$
        \ENDIF
        \STATE Compute $\widehat G_t=B_s^\top G_t$
        \STATE Update the compact momentum
        \[
        H_t=
        \begin{cases}
        (1-\beta)B_s^\top B_{s-1}H_{t-1}+\beta\widehat G_t,
        & t=s\tau,\ s>0,\\
        (1-\beta)H_{t-1}+\beta\widehat G_t,
        & \text{otherwise}.
        \end{cases}
        \]
        \STATE Update $X_{t+1}=X_t-\eta B_s\,\msign(H_t)$
    \ENDFOR
\ENDFOR
\RETURN $X_T$.
\end{algorithmic}
\end{algorithm}

\subsection{Robustness Across Total Ranks and Source Trajectories}\label{app:stagewise-rank-details}
\begin{figure}[t]
    \centering
    \begin{subfigure}[t]{0.48\linewidth}
        \centering
        \includegraphics[width=\linewidth]{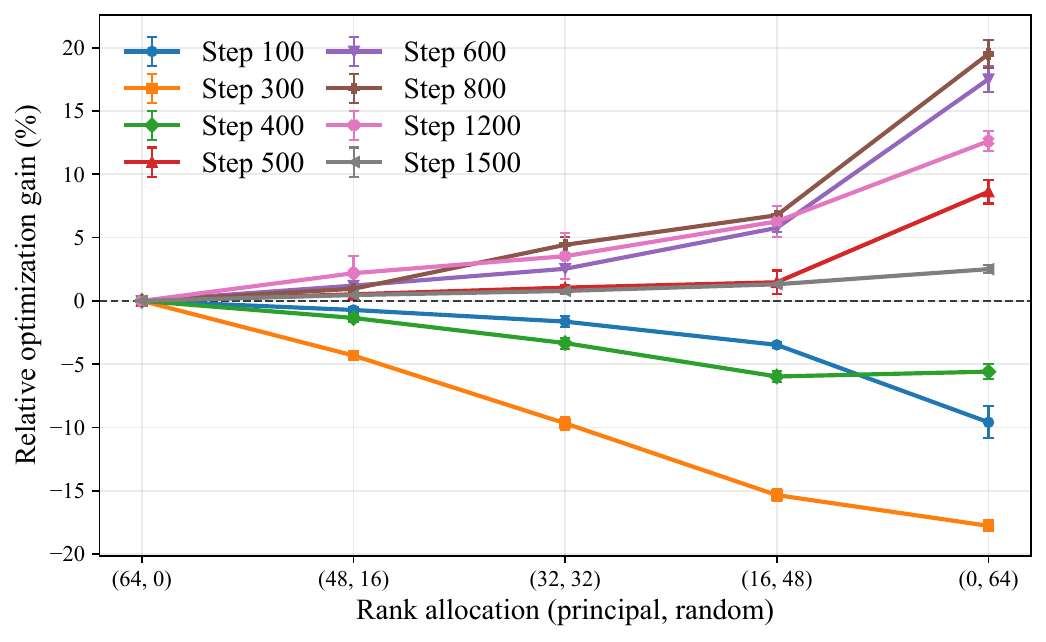}
        \caption{Full-Muon, rank 64.}
    \end{subfigure}
    \hfill
    \begin{subfigure}[t]{0.48\linewidth}
        \centering
        \includegraphics[width=\linewidth]{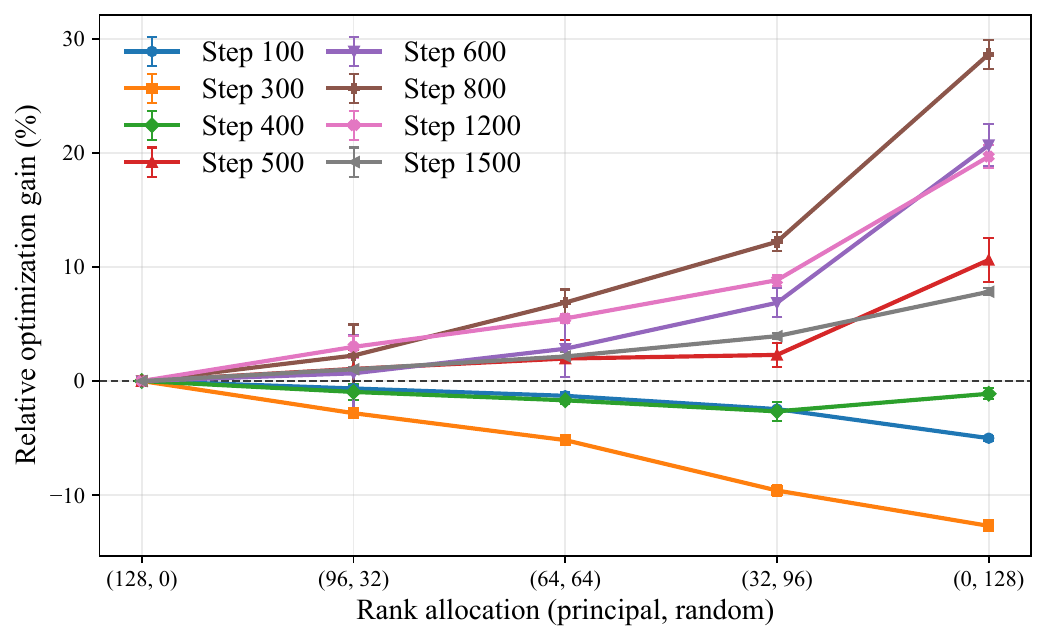}
        \caption{Full-Muon, rank 128.}
    \end{subfigure}
    \begin{subfigure}[t]{0.48\linewidth}
        \centering
        \includegraphics[width=\linewidth]{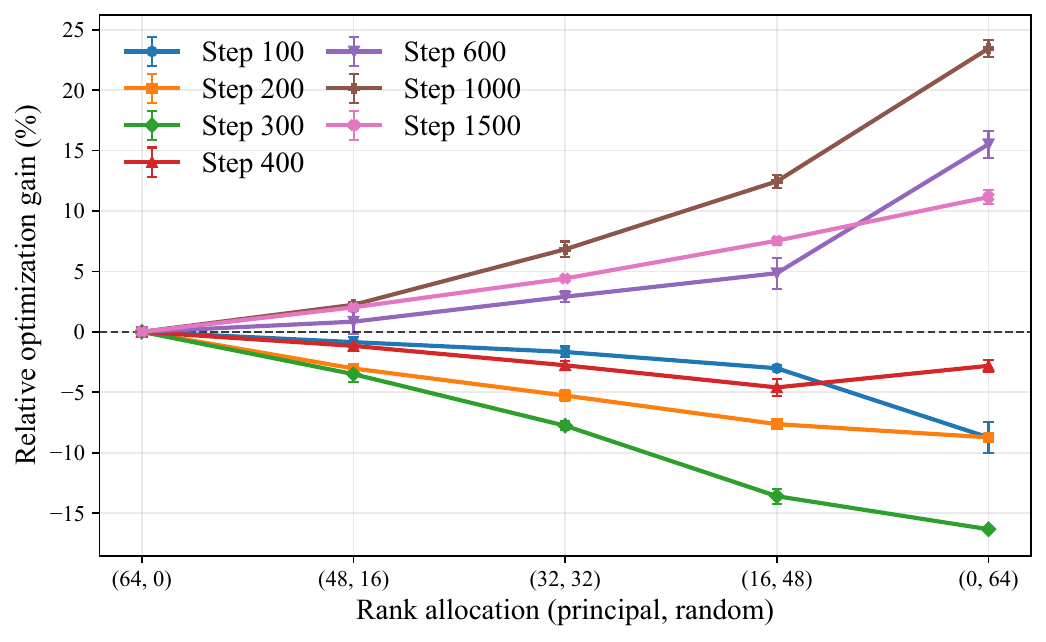}
        \caption{GaLore-Muon, rank 64.}
    \end{subfigure}
    \hfill
    \begin{subfigure}[t]{0.48\linewidth}
        \centering
        \includegraphics[width=\linewidth]{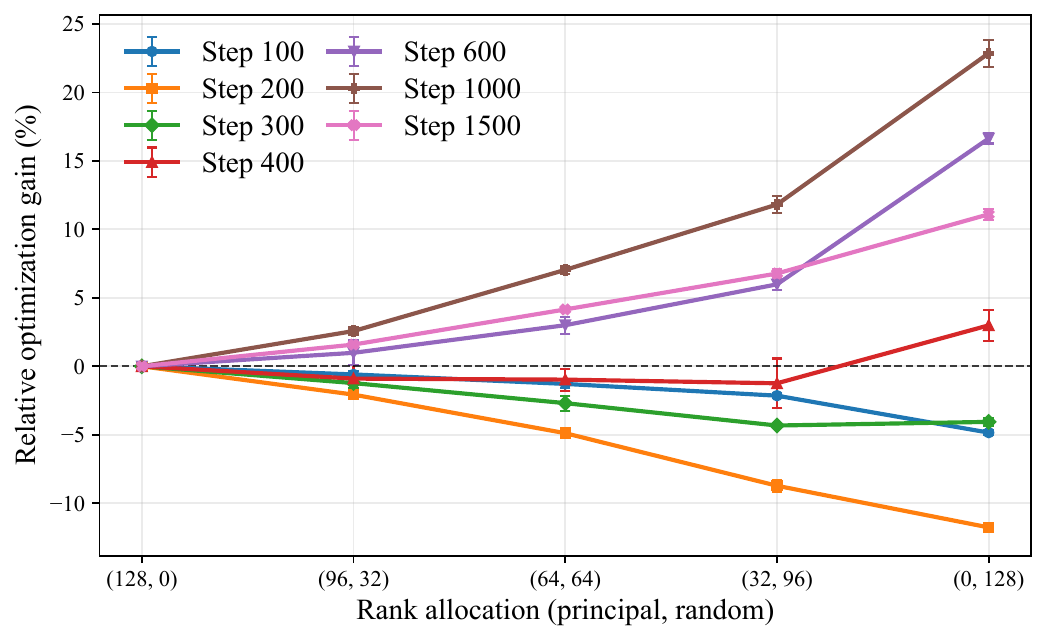}
        \caption{GaLore-Muon, rank 128.}
    \end{subfigure}
    \caption{\textbf{Stage-wise rank-allocation preference across total ranks and source trajectories.}
    Relative gain in 200-step validation-loss reduction over the principal-only allocation across principal-random rank allocations. Results are averaged over three independent seeds.}
    \label{fig:rank-allocation-rank-robustness}
\end{figure}

We further evaluate the robustness of the stage-wise rank-allocation preference across total ranks and source trajectories. In addition to rank $256$ in Fig.~\ref{fig:subspace-rank-transition}(b)--(c), we repeat the 200-step interventions along Full-Muon and GaLore-Muon trajectories with total ranks $k=r+q\in\{64,128\}$. For each rank, we evaluate random-rank fractions $q/k\in\{0,0.25,0.5,0.75,1\}$ over three independent seeds.
All branches use a Muon learning rate of $10^{-2}$ and a learning rate of $5\times10^{-3}$ for the parameters updated by AdamW. The results are shown in Fig.~\ref{fig:rank-allocation-rank-robustness}.
Across ranks $64$, $128$, and $256$ and both source trajectories, the preference consistently shifts from principal-heavy allocations early in training toward random-heavy allocations later.

\begin{table}[t]
\centering
\small
\caption{Learning rates for LLaMA pre-training on C4. For Muon-based methods, entries report the Muon learning rate; AdamW-updated parameters use $5\times10^{-3}$, $3\times10^{-3}$, and $2\times10^{-3}$ for LLaMA-60M, 130M, and 350M, respectively.}
\label{tab:pretrain-hyperparameters}
\setlength{\tabcolsep}{2.5pt}
\begin{tabular}{lccccccc}
\toprule
Model
& AdamW
& Full-Muon
& GaLore-Muon
& GoLore-Muon
& GUM
& \dramf
& \dram \\
\midrule
LLaMA-60M
& $5\!\times\!10^{-3}$
& $1\!\times\!10^{-2}$
& $1\!\times\!10^{-2}$
& $1\!\times\!10^{-2}$
& $1\!\times\!10^{-2}$
& $1.2\!\times\!10^{-2}$
& $8\!\times\!10^{-3}$ \\
LLaMA-130M
& $3\!\times\!10^{-3}$
& $1\!\times\!10^{-2}$
& $1\!\times\!10^{-2}$
& $6\!\times\!10^{-3}$
& $6\!\times\!10^{-3}$
& $1\!\times\!10^{-2}$
& $1\!\times\!10^{-2}$ \\
LLaMA-350M
& $2\!\times\!10^{-3}$
& $6\!\times\!10^{-3}$
& $8\!\times\!10^{-3}$
& $6\!\times\!10^{-3}$
& $8\!\times\!10^{-3}$
& $8\!\times\!10^{-3}$
& $8\!\times\!10^{-3}$ \\
\bottomrule
\end{tabular}
\end{table}

\begin{table}[t]
\centering
\small
\caption{Projection ranks for pre-training experiments. For GUM, $p$ is the full-rank correction probability. For \dram, $(r,q)$ denotes the principal and random ranks, with \dramf fixed at $(128,128)$ and \dram linearly reallocating from $(255,1)$ to $(0,256)$.}
\label{tab:projection-configs}
\setlength{\tabcolsep}{5pt}
\begin{tabular}{lccccc}
\toprule
Model
& GaLore-Muon
& GoLore-Muon
& GUM
& \dramf
& \dram\\
\midrule
LLaMA-60M
& 256
& 256
& 128 ($p=0.53$)
& $(128,128)$
& $(255,1)\!\rightarrow\!(0,256)$ \\
LLaMA-130M
& 256
& 256
& 128 ($p=0.32$)
& $(128,128)$
& $(255,1)\!\rightarrow\!(0,256)$ \\
LLaMA-350M
& 256
& 256
& 128 ($p=0.23$)
& $(128,128)$
& $(255,1)\!\rightarrow\!(0,256)$ \\
\bottomrule
\end{tabular}
\end{table}

\subsection{Pre-training Hyperparameters} \label{app:pretraining-hyperparameters}
All models are trained from scratch on C4 with sequence length $1024$, global batch size $512$, and bfloat16 precision. To improve the reliability of the results, we use scale-specific search grids and report the best-performing configuration for each method.
For Muon-based methods, we search the Muon learning rate over $\{0.8, 1.0, 1.2\}\times 10^{-2}$ for LLaMA-60M, $\{0.6, 0.8, 1.0\}\times 10^{-2}$ for LLaMA-130M, and $\{4, 6, 8\}\times 10^{-3}$ for LLaMA-350M.
The selected learning rates are summarized in Table~\ref{tab:pretrain-hyperparameters}.
For projection-based methods, we refresh the projection every 50 steps for LLaMA-60M and every 100 steps for LLaMA-130M and LLaMA-350M, with projection configurations shown in Table~\ref{tab:projection-configs}.
For GUM, we fix the projection rank at $128$ and choose the full-rank update probability $p$ so that its expected memory cost approximately matches that of the other memory-efficient methods.
\begin{table}[t]
\centering
\small
\setlength{\tabcolsep}{5pt}
\caption{Learning-rate search spaces for the fine-tuning experiments.}
\label{tab:finetune-lr-pools}
\begin{tabular}{llcc}
\toprule
Task & Backbone
& $\eta_{\mathrm{Muon}}$
& $\eta_{\mathrm{AdamW}}$ \\
\midrule
IFEval
& All
& $\{1,3,5,8,10,15,20,25,30\}\times10^{-5}$
& $3\times10^{-5}$ \\
\midrule
GSM8K
& Qwen3-8B
& $\{1,3,5,8,10\}\times10^{-5}$
& $\{1,3\}\times10^{-5}$ \\
GSM8K
& \makecell[l]{LLaMA-3.1-8B,\\Gemma-2-9B}
& $\{1,3,5,8,10,15,20,25,30\}\times10^{-5}$
& $\{1,3\}\times10^{-5}$ \\
\midrule
MBPP
& Qwen3-8B
& $\{1,5,10,30,50,80,100\}\times10^{-7}$
& $\{1,5,10\}\times10^{-7}$ \\
MBPP
& \makecell[l]{LLaMA-3.1-8B,\\Gemma-2-9B}
& $\{1,1.5,2,2.5,3\}\times10^{-4}$
& $\{1,3\}\times10^{-5}$ \\
\bottomrule
\end{tabular}
\end{table}

\begin{table}[t]
\centering
\small
\setlength{\tabcolsep}{5pt}
\caption{
Selected learning-rate pairs $(\eta_{\mathrm{Muon}},\eta_{\mathrm{AdamW}})$
for Table~\ref{tab:fine_tune}, in units of $10^{-5}$.
}
\label{tab:finetune-selected-lrs}
\begin{tabular}{llccccc}
\toprule
Task & Backbone
& Full-Muon
& GaLore-Muon
& GoLore-Muon
& GUM
& \dram \\
\midrule

\multirow{3}{*}{IFEval}
& Qwen3-8B
& $(25,3)$
& $(25,3)$
& $(30,3)$
& $(25,3)$
& $(30,3)$ \\

& LLaMA-3.1-8B
& $(30,3)$
& $(30,3)$
& $(25,3)$
& $(25,3)$
& $(25,3)$ \\

& Gemma-2-9B
& $(10,3)$
& $(15,3)$
& $(25,3)$
& $(15,3)$
& $(15,3)$ \\

\midrule

\multirow{3}{*}{GSM8K}
& Qwen3-8B
& $(8,1)$
& $(10,1)$
& $(10,1)$
& $(8,3)$
& $(10,1)$ \\

& LLaMA-3.1-8B
& $(8,3)$
& $(10,1)$
& $(10,3)$
& $(8,3)$
& $(10,3)$ \\

& Gemma-2-9B
& $(15,3)$
& $(30,1)$
& $(30,3)$
& $(30,3)$
& $(25,1)$ \\

\midrule

\multirow{3}{*}{MBPP}
& Qwen3-8B
& $(0.5,0.1)$
& $(0.05,0.1)$
& $(0.05,0.1)$
& $(0.01,0.1)$
& $(0.1,0.1)$ \\

& LLaMA-3.1-8B
& $(10,3)$
& $(10,3)$
& $(25,1)$
& $(30,1)$
& $(30,1)$ \\

& Gemma-2-9B
& $(10,1)$
& $(15,3)$
& $(15,3)$
& $(20,1)$
& $(15,3)$ \\

\bottomrule
\end{tabular}
\end{table}

\subsection{Fine-tuning Hyperparameters}
\label{app:finetuning-hyperparameters}
For GPT4-LLM-Cleaned, we use the version released by
teknium\footnote{https://huggingface.co/datasets/teknium/GPT4-LLM-Cleaned}.
For each task--backbone pair, we tune $\eta_{\mathrm{Muon}}$ for the weights optimized by Muon and $\eta_{\mathrm{AdamW}}$ for the remaining parameters updated by AdamW. All methods use the same learning-rate grid within each comparison. Table~\ref{tab:finetune-lr-pools} reports the search spaces, and Table~\ref{tab:finetune-selected-lrs} reports the selected learning-rate pairs corresponding to Table~\ref{tab:fine_tune}. Projection bases are refreshed every $100$ steps for IFEval and every $50$ steps for GSM8K and MBPP.
For GUM, we use rank 128 with full-rank correction probabilities $p=0.156$, $0.149$, and $0.162$ for Qwen3-8B, LLaMA-3.1-8B, and Gemma-2-9B, respectively, chosen to approximately match the expected optimizer-state storage of the rank-512 projection methods.

For instruction following, we fine-tune on 54,546 GPT4-LLM-Cleaned examples with a global batch size of $16$ and evaluate on the official 541 IFEval prompts. For mathematical reasoning, we fine-tune on 7,473 GSM8K examples with a global batch size of $16$ and evaluate on the official 1,319-example test set. For code generation, we fine-tune on 374 MBPP tasks with a global batch size of $1$ and evaluate on the official 500-task test set. All experiments use one epoch, bfloat16 precision, and cosine learning-rate decay.

\subsection{GPU Memory Usage}\label{app:memory}
We report the peak GPU memory allocated during full-parameter fine-tuning in Table~\ref{tab:ft-memory}.
\dram reduces peak GPU memory by 17.9--20.7\% relative to Full-Muon
across the nine fine-tuning settings.
Its peak memory differs from that of rank-matched GaLore-Muon and GoLore-Muon by at most $0.04$ GiB across the nine settings,
indicating negligible additional memory overhead from the principal-random construction and stage-wise rank reallocation.
GUM exhibits slightly higher peak memory across all settings, consistent with its probabilistic full-rank corrections.

\begin{table}[t]
\centering
\small
\caption{
Peak allocated GPU memory (GiB) during full-parameter fine-tuning.
}
\label{tab:ft-memory}
\setlength{\tabcolsep}{5pt}
\begin{tabular}{llccccc}
\toprule
Task & Backbone
& Full-Muon
& GaLore-Muon
& GoLore-Muon
& GUM
& \dram \\
\midrule
\multirow{3}{*}{IFEval}
& Qwen3-8B       & 57.70 & 47.38 & 47.37 & 47.45 & 47.38 \\
& LLaMA-3.1-8B   & 50.86 & 40.36 & 40.36 & 41.31 & 40.35 \\
& Gemma-2-9B     & 60.25 & 48.04 & 48.08 & 48.83 & 48.07 \\
\midrule
\multirow{3}{*}{GSM8K}
& Qwen3-8B       & 52.86 & 42.52 & 42.51 & 42.83 & 42.51 \\
& LLaMA-3.1-8B   & 50.86 & 40.36 & 40.35 & 40.73 & 40.35 \\
& Gemma-2-9B     & 61.45 & 49.24 & 49.29 & 49.67 & 49.28 \\
\midrule
\multirow{3}{*}{MBPP}
& Qwen3-8B       & 52.80 & 42.47 & 42.47 & 43.23 & 42.47 \\
& LLaMA-3.1-8B   & 50.81 & 40.31 & 40.31 & 40.71 & 40.31 \\
& Gemma-2-9B     & 60.25 & 48.03 & 48.08 & 48.45 & 48.06 \\
\bottomrule
\end{tabular}
\end{table}

\section{Detailed Convergence Analysis}
\label{app:dynamic_convergence}
This appendix provides the bound underlying Theorem~\ref{thm:dram_convergence} and Corollary~\ref{cor:dram_fixed_allocation}, together with their proofs.

Since the target horizon $T$ need not be divisible by the block length $\tau$, the final block of the actual run may contain fewer than $\tau$ updates. We therefore set $S=\lceil T/\tau\rceil$ and run Algorithm~\ref{alg:DRAM} for exactly $T$ updates. To reuse the complete-block analysis below, we introduce the auxiliary horizon $\widetilde T:=S\tau$ and consider a conceptual continuation of the final block from $T$ to $\widetilde T$.
Throughout this appendix, the rank-allocation schedule is prescribed independently of the optimization randomness. For the complete-block extension, define $\mathcal I_s:=\{s\tau,\ldots,(s+1)\tau-1\}$, $k_s=r_s+q_s$, $\Delta_s:=\E[F(X_{s\tau})-F(X_{(s+1)\tau})]$ for $s=0,\ldots,S-1$, and $\Delta=F(X_0)-F^*$.

For the stochastic setting, let $\delta_s=q_s/(m-r_s)$, where $0\le r_s<m$ and $1\le q_s\le m-r_s.$ For the deterministic setting, we allow $0\le r_s\le m$, $0\le q_s\le m-r_s$, $k_s>0$ and define $\gamma_s=k_s/m$.

\subsection{Finite-Time Bound and Parameter Choices} \label{app:finite_time_bound}
\begin{theorem} \label{thm:finite_time_bound}
Let $\widetilde T=S\tau$ and $c_\beta:=1-\frac{2(1-\beta)}{\beta\tau}$. Suppose Assumptions~\ref{ass:lower_bound}--\ref{ass:exact_muon} hold and $c_\beta>0$. For $\sigma>0$, assume \(0\le r_s<m\), \(1\le q_s\le m-r_s\) for all \(s\), and
\(\delta_{s+1}\ge \delta_s\) for \(s=0,\ldots,S-2\).
Then the conceptual complete-block extension of Algorithm~\ref{alg:DRAM} satisfies
\[
\begin{aligned}
\frac1{\widetilde T}\sum_{t=0}^{\widetilde T-1} \E\left[\|\nabla F(X_t)\|_*\right]
\le& \frac{\Delta}{\eta c_\beta \widetilde T\delta_0} + \left[ \frac{4(1-\beta)(\tau-1)L_*\eta}{\beta\tau c_\beta} + \frac{L_*\eta}{2c_\beta} \right] \frac1S \sum_{s=0}^{S-1} \frac{1}{\delta_s} \\
&+ \frac{2\sigma}{c_\beta} \sqrt{\frac{\beta}{2-\beta}} \cdot \frac1S \sum_{s=0}^{S-1} \frac{\sqrt{k_s}}{\delta_s} + \frac{L_*\eta(\tau-1)}{2S} \sum_{s=0}^{S-1} \frac{1+\delta_s}{\delta_s}.
\end{aligned}
\]
For $\sigma=0$, the rank conditions can instead be relaxed to
$0\le r_s\le m$, $0\le q_s\le m-r_s$, and
$k_s=r_s+q_s>0$ for all $s$. If $\gamma_{s+1}\ge\gamma_s$ for $s=0,\ldots,S-2$, then the conceptual complete-block extension satisfies
\[
\begin{aligned}
\frac1{\widetilde T}\sum_{t=0}^{\widetilde T-1}
\E\left[\|\nabla F(X_t)\|_*\right]
\le&
\frac{\Delta}{\eta c_\beta \widetilde T\gamma_0}
+
\left[
\frac{4(1-\beta)(\tau-1)L_*\eta}{\beta\tau c_\beta}
+\frac{L_*\eta}{2c_\beta}
\right]
\frac1S\sum_{s=0}^{S-1}\frac{1}{\gamma_s}
\\
&+
\frac{L_*\eta(\tau-1)}{2S}
\sum_{s=0}^{S-1}\frac{1+\gamma_s}{\gamma_s}.
\end{aligned}
\]
\end{theorem}
The proof combines compact Muon descent, blockwise momentum tracking, and a projected-to-full-gradient conversion.

\paragraph{Parameter choices for Theorem~\ref{thm:dram_convergence}.}
Consider the nontrivial regime $L_*\Delta>0$ and fix $\beta_0\in(0,1]$ independently of $T$. For $\sigma>0$, fix an a priori rank cap $1\le k_{\max}\le m$ and assume \[T\ge\max\left\{\left\lceil\frac{3}{\beta_0}\right\rceil,\left\lceil\frac{9k_{\max}\sigma^2}{L_*\Delta}\right\rceil\right\}.\]
Set \[\beta=\min\left\{\beta_0,\frac{\sqrt{L_*\Delta}}{\sqrt{k_{\max}}\sigma\sqrt{T}}\right\},\quad \tau=\left\lceil\frac{3}{\beta}\right\rceil,\quad S=\left\lceil\frac{T}{\tau}\right\rceil,\quad \widetilde T=S\tau,\quad \eta=\sqrt{\frac{\Delta\beta}{5L_*\widetilde T}}.\] After $S$ is determined, let $\{(r_s,q_s)\}_{s=0}^{S-1}$ be any prescribed rank-allocation schedule fixed independently of the optimization randomness and satisfying the stochastic conditions of Theorem~\ref{thm:dram_convergence}, including $k_s\le k_{\max}$. Thus, the schedule does not affect the choice of $\beta$, $\tau$, or $S$. The horizon condition ensures $\beta\ge3/T$: if $\beta=\beta_0$, this follows from $T\ge3/\beta_0$; otherwise, $T\ge9k_{\max}\sigma^2/(L_*\Delta)$ gives $\sqrt{L_*\Delta}/(\sqrt{k_{\max}}\sigma\sqrt{T})\ge3/T$. Hence $\tau\le T$, so $T\le\widetilde T<T+\tau\le2T$. Moreover, $\beta\tau\ge3$, $\tau-1<3/\beta$, and $c_\beta\ge1/3$.

For $\sigma=0$, assume $T\ge\lceil3/\beta_0\rceil$ and set \[\beta=\beta_0,\qquad \tau=\left\lceil\frac{3}{\beta_0}\right\rceil,\qquad S=\left\lceil\frac{T}{\tau}\right\rceil,\qquad \widetilde T=S\tau,\qquad \eta=\sqrt{\frac{\Delta\beta_0}{5L_*\widetilde T}}.\] After $S$ is determined, let $\{(r_s,q_s)\}_{s=0}^{S-1}$ be any prescribed rank-allocation schedule satisfying the deterministic conditions of Theorem~\ref{thm:dram_convergence}. Here $\tau\le T$, $T\le\widetilde T<T+\tau\le2T$, and $c_\beta\ge1/3$.

\subsection{Supporting Lemmas} \label{app:supporting_lemmas}
\begin{lemma}[One-step descent for \dram]\label{lem:space_muon_one_step_descent}
Suppose Assumptions~\ref{ass:nuclear_smoothness} and~\ref{ass:exact_muon} hold. Fix a block $s$ and an iteration $t\in\mathcal I_s$. Let $B_s=[R_s\ P_s]\in\mathbb{R}^{m\times k_s}$ satisfy $B_s^\top B_s=I_{k_s}$ and let \(H_t\in\mathbb{R}^{k_s\times n}\). If $X_{t+1}=X_t-\eta B_s\,\msign(H_t)$, then
\[
F(X_{t+1})
\le F(X_t)-\eta\|B_s^\top\nabla F(X_t)\|_*+2\eta\|H_t-B_s^\top\nabla F(X_t)\|_*+\frac{L_*\eta^2}{2}.
\]
\end{lemma}

\begin{proof}
Write $A_t=B_s^\top\nabla F(X_t)$ and $Q_t=\msign(H_t)$.
Since $B_s^\top B_s=I$ and $\|Q_t\|_{\op}\le1$, nuclear smoothness gives
\[
F(X_{t+1})\le F(X_t)-\eta\langle A_t,Q_t\rangle_F+\frac{L_*\eta^2}{2}.
\]
By exact matrix-sign normalization and nuclear/operator norm duality,
\begin{align*}
\langle A_t,Q_t\rangle_F
&=\|H_t\|_*+\langle A_t-H_t,Q_t\rangle_F\\
&\ge\|H_t\|_*-\|A_t-H_t\|_*\\
&\ge\|A_t\|_*-2\|H_t-A_t\|_*.
\end{align*}
Substitution proves the stated descent inequality directly in compact coordinates.
\end{proof}

\begin{lemma}[Blockwise compact Muon tracking]\label{lem:space_muon_tracking}
Suppose Assumptions~\ref{ass:gradient_filtration}--\ref{ass:exact_muon} hold.
Fix a block \(s\), and let \(B_s=[R_s\ P_s]\) be the blockwise basis used by the conceptual complete-block extension of Algorithm~\ref{alg:DRAM}.
For \(j=0,\ldots,\tau-1\), define
\[
A_j:=B_s^\top\nabla F(X_{s\tau+j}), \qquad \widehat G_j:=B_s^\top G_{s\tau+j}, \qquad H_j:=H_{s\tau+j}.
\]
Let $\lambda:=1-\beta$. Then
\[
\sum_{j=0}^{\tau-1}\E\left[\|H_j-A_j\|_*\right]
\le \frac{1-\beta}{\beta\tau}\sum_{j=0}^{\tau-1}\E\left[\|A_j\|_*\right]+\frac{2(1-\beta)}{\beta}(\tau-1)L_*\eta+\tau\sqrt{k_s}\sigma
\sqrt{\frac{\beta}{2-\beta}}.
\]
\end{lemma}

\begin{proof}
Within block \(s\), the basis \(B_s\) is fixed after the block start. By Algorithm~\ref{alg:DRAM},
\[
H_0=\beta\widehat G_0,
\qquad
H_j=(1-\beta)H_{j-1}+\beta\widehat G_j,\quad j\ge 1.
\]
Let $E_j:=H_j-A_j$ and $\zeta_j:=\widehat G_j-A_j=B_s^\top\left(G_{s\tau+j}-\nabla F(X_{s\tau+j})\right)$. Then $E_0=-\lambda A_0+\beta\zeta_0$, and for \(j\ge1\),
\[
\begin{aligned}
E_j
&=(1-\beta)H_{j-1}+\beta\widehat G_j-A_j  \\
&=\lambda E_{j-1}+\lambda(A_{j-1}-A_j)+\beta\zeta_j .
\end{aligned}
\]
Unrolling this recursion gives
\[
E_j
=-\lambda^{j+1}A_0
+\sum_{i=1}^{j}\lambda^{j-i+1}(A_{i-1}-A_i)
+\beta\sum_{i=0}^{j}\lambda^{j-i}\zeta_i.
\]
Taking nuclear norms, using the triangle inequality and summing over $j=0,\cdots,\tau-1$, we obtain
\[
\begin{aligned}
\sum_{j=0}^{\tau-1}\|E_j\|_*
\le
\sum_{j=0}^{\tau-1}\lambda^{j+1}\|A_0\|_*
+\sum_{j=0}^{\tau-1}\sum_{i=1}^{j}\lambda^{j-i+1}\|A_{i-1}-A_i\|_*+\sum_{j=0}^{\tau-1}\left\|\beta\sum_{i=0}^{j}\lambda^{j-i}\zeta_i\right\|_*.
\end{aligned}
\]
We control the three terms separately.

For the initial term, summing the geometric series yields
\[
\sum_{j=0}^{\tau-1}\lambda^{j+1}\|A_0\|_*
\le
\frac{\lambda}{\beta}\|A_0\|_*.
\]
To relate \(\|A_0\|_*\) to the average projected gradient over the block, we use telescoping: for every \(j\),
\[
\|A_0\|_*
\le
\|A_j\|_*+\sum_{i=1}^{j}\|A_{i-1}-A_i\|_*.
\]
Averaging this inequality over \(j=0,\ldots,\tau-1\) gives
\[
\|A_0\|_*
\le\frac1\tau\sum_{j=0}^{\tau-1}\|A_j\|_*+\sum_{i=1}^{\tau-1}\|A_i-A_{i-1}\|_*.
\]
Next, we control the variation of \(A_j\) within the block. By the definition of \(A_j\), we obtain
\[
\begin{aligned}
\|A_i-A_{i-1}\|_*
&=
\|B_s^\top(
\nabla F(X_{s\tau+i})-\nabla F(X_{s\tau+i-1}))\|_*\\
&\le
\|
\nabla F(X_{s\tau+i})-\nabla F(X_{s\tau+i-1})\|_*\\
&\le
L_*\|X_{s\tau+i}-X_{s\tau+i-1}\|_{\op}
\le L_*\eta.
\end{aligned}
\]
Substituting this drift bound into the previous estimate gives
\[
\sum_{j=0}^{\tau-1}\lambda^{j+1}\|A_0\|_*
\le
\frac{\lambda}{\beta\tau}
\sum_{j=0}^{\tau-1}\|A_j\|_*
+
\frac{\lambda}{\beta}(\tau-1)L_*\eta .
\]

Second, we bound the accumulated drift term. Using again
\(\|A_i-A_{i-1}\|_*\le L_*\eta\), we obtain
\[
\begin{aligned}
\sum_{j=0}^{\tau-1}
\sum_{i=1}^{j}\lambda^{j-i+1}\|A_{i-1}-A_i\|_*
\le
\sum_{j=1}^{\tau-1}
\sum_{i=1}^{j}\lambda^{j-i+1}L_*\eta
\le
\sum_{j=1}^{\tau-1}\frac{\lambda}{\beta}L_*\eta
\le
\frac{\lambda}{\beta}(\tau-1)L_*\eta.
\end{aligned}
\]

Third, we control the stochastic noise term. Since the matrix inside the nuclear norm lies in
\(\mathbb{R}^{k_s\times n}\), its rank is at most \(k_s\). Hence
\[
\E\left\|\beta\sum_{i=0}^{j}\lambda^{j-i}\zeta_i\right\|_*
\le
\sqrt{k_s} \ \E\left\|\beta\sum_{i=0}^{j}\lambda^{j-i}\zeta_i\right\|_F
\le
\sqrt{k_s}\left(\E\left\|\beta\sum_{i=0}^{j}\lambda^{j-i}\zeta_i\right\|_F^2\right)^{1/2}.
\]
We now bound the Frobenius second moment.

For $0\le i<f\le j$, note that $f\ge1$. By iteration $s\tau+f$, the block basis $B_s$ has already been constructed, and both $B_s$ and $\zeta_i$ are $\mathcal F_{s\tau+f}$-measurable. Moreover, by Assumption~\ref{ass:gradient_filtration},
\[
\E[\zeta_f\mid\mathcal F_{s\tau+f}]
=
B_s^\top
\E\!\left[
G_{s\tau+f}-\nabla F(X_{s\tau+f})
\mid\mathcal F_{s\tau+f}
\right]
=0.
\]
Hence, by the tower property, we obtain $\E\langle\zeta_i,\zeta_f\rangle_F
=\E\!\left[\left\langle\zeta_i,\E[\zeta_f\mid\mathcal F_{s\tau+f}]\right\rangle_F\right]=0.$
Therefore,
\[
\E\left[\left\|\beta\sum_{i=0}^j\lambda^{j-i}\zeta_i\right\|_F^2\right]
=
\beta^2\sum_{i=0}^j\lambda^{2(j-i)}
\E\|\zeta_i\|_F^2.
\]

It remains to bound the individual second moments. Although $B_s$ depends on the block-start stochastic gradient, no independence is needed here. Since $B_s^\top B_s=I$, the projection is contractive in Frobenius norm, and hence
\[
\|\zeta_i\|_F=\left\|B_s^\top\bigl(G_{s\tau+i}-\nabla F(X_{s\tau+i})\bigr)\right\|_F
\le\|G_{s\tau+i}-\nabla F(X_{s\tau+i})\|_F.
\]
Taking expectations and applying Assumption~\ref{ass:gradient_filtration} gives $\E\|\zeta_i\|_F^2\le\sigma^2$. Consequently,
\[
\E\left[\left\|\beta\sum_{i=0}^j\lambda^{j-i}\zeta_i\right\|_F^2\right]
\le\beta^2\sigma^2\sum_{i=0}^j\lambda^{2(j-i)}
\le\frac{\beta}{2-\beta}\sigma^2,
\]
where we used $1-\lambda^2=1-(1-\beta)^2=\beta(2-\beta).$
Finally, since the matrix inside the norm has at most $k_s$ rows,
\[
\E\left[\left\|\beta\sum_{i=0}^j\lambda^{j-i}\zeta_i\right\|_*\right]
\le\sqrt{k_s}\left(\E\left[\left\|\beta\sum_{i=0}^j\lambda^{j-i}\zeta_i\right\|_F^2\right]\right)^{1/2}
\le\sqrt{k_s}\sigma\sqrt{\frac{\beta}{2-\beta}}.
\]

Combining the bounds for the initial-error term, drift, and noise terms, we get
\[
\sum_{j=0}^{\tau-1}\E\left[\|E_j\|_*\right]
\le\frac{\lambda}{\beta\tau}\sum_{j=0}^{\tau-1}\E\left[\|A_j\|_*\right]+\frac{2\lambda}{\beta}(\tau-1)L_*\eta+\tau\sqrt{k_s}\sigma\sqrt{\frac{\beta}{2-\beta}} .
\]
Finally, substituting \(\lambda=1-\beta\) proves the claim.
\end{proof}

\begin{lemma}[Weighted projected stationarity]\label{lem:space_muon_projected}
Suppose Assumptions~\ref{ass:lower_bound}--\ref{ass:exact_muon} hold.
Assume $0\le r_s<m$ and $1\le q_s\le m-r_s$ for all $s$, and define
$\delta_s=q_s/(m-r_s)$.
Let $\widetilde T=S\tau$, $\Delta=F(X_0)-F^*$, and
$c_\beta=1-\frac{2(1-\beta)}{\beta\tau}$.
If \(c_\beta>0\), then the iterates in the conceptual complete-block extension of Algorithm~\ref{alg:DRAM} satisfy
\begin{equation} \label{eqn:lem:projection_bound}
\begin{split}
&\frac1{\widetilde T} \sum_{s=0}^{S-1} \frac{1}{\delta_s}\sum_{j=0}^{\tau-1}\E\left[\|B_s^\top\nabla F(X_{s\tau+j})\|_*\right] \\
\le& \frac{1}{\eta c_\beta \widetilde T} \sum_{s=0}^{S-1} \frac{\Delta_s}{\delta_s} + \left[ \frac{4(1-\beta)(\tau-1)L_*\eta}{\beta\tau c_\beta} + \frac{L_*\eta}{2c_\beta} \right] \frac1S \sum_{s=0}^{S-1} \frac{1}{\delta_s} + \frac{2\sigma}{c_\beta} \sqrt{\frac{\beta}{2-\beta}} \cdot \frac1S \sum_{s=0}^{S-1} \frac{\sqrt{k_s}}{\delta_s}.
\end{split}
\end{equation}
\end{lemma}

\begin{proof}
Fix a block \(s\). To simplify notation, define the accumulated projected gradient norm in this block as $S_s:=\sum_{j=0}^{\tau-1}\E\left[\|B_s^\top\nabla F(X_{s\tau+j})\|_*\right]$, and define the blockwise function decrease as $\Delta_s=\E\left[F(X_{s\tau})-F(X_{(s+1)\tau})\right]$.

We first apply the one-step descent inequality in Lemma~\ref{lem:space_muon_one_step_descent} to all
iterations in block \(s\). Summing over \(j=0,\ldots,\tau-1\) and taking expectation gives
\[
\eta S_s
\le\Delta_s+2\eta\sum_{j=0}^{\tau-1}\E\bigl[\|H_{s\tau+j}-B_s^\top\nabla F(X_{s\tau+j})\|_*\bigr]+\frac{\tau L_*\eta^2}{2}.
\]
Using Lemma~\ref{lem:space_muon_tracking}, we obtain
\[
\begin{aligned}
\eta S_s
\le&
\Delta_s
+2\eta\left[
\frac{1-\beta}{\beta\tau}S_s
+\frac{2(1-\beta)}{\beta}(\tau-1)L_*\eta
+\tau\sqrt{k_s}\sigma\sqrt{\frac{\beta}{2-\beta}}
\right]
+\frac{\tau L_*\eta^2}{2}.
\end{aligned}
\]
Moving the $S_s$ term to the left and dividing by $\eta c_\beta$ gives the per-block estimate
\begin{equation} \label{eqn:per_block_Ss}
S_s
\le
\frac{\Delta_s}{\eta c_\beta}
+\frac{4(1-\beta)(\tau-1)L_*\eta}{\beta c_\beta}
+\frac{2\tau\sqrt{k_s}\sigma}{c_\beta}\sqrt{\frac{\beta}{2-\beta}}
+\frac{\tau L_*\eta}{2c_\beta},
\end{equation}
where $c_\beta=1-\frac{2(1-\beta)}{\beta\tau}$.
Therefore,
\begin{align*}
\frac1{\widetilde T} \sum_{s=0}^{S-1} \frac{1}{\delta_s}S_s
\le& \frac{1}{\eta c_\beta \widetilde T} \sum_{s=0}^{S-1} \frac{\Delta_s}{\delta_s} + \frac{4(1-\beta)(\tau-1)L_*\eta}{\beta c_\beta \widetilde T} \sum_{s=0}^{S-1} \frac{1}{\delta_s} \\
&+ \frac{2\tau\sigma}{c_\beta \widetilde T} \sqrt{\frac{\beta}{2-\beta}} \sum_{s=0}^{S-1} \frac{\sqrt{k_s}}{\delta_s} + \frac{\tau L_*\eta}{2c_\beta \widetilde T} \sum_{s=0}^{S-1} \frac{1}{\delta_s}. \end{align*}
Using \(\widetilde T=S\tau\) gives~(\ref{eqn:lem:projection_bound}), which completes the proof.
\end{proof}

\begin{lemma}[Coverage for projected nuclear stationarity]
\label{lem:projected_nuclear_coverage}
Fix a block \(s\in\{0,\ldots,S-1\}\). Let $\Pi_s:=R_sR_s^\top$, $Q_s:=I-\Pi_s$ and $B_s=[R_s\ P_s]$, where \(R_s^\top R_s=I_{r_s}\) and \(P_s\) is sampled as in Algorithm~\ref{alg:DRAM}.
Suppose Assumptions~\ref{ass:nuclear_smoothness} and~\ref{ass:exact_muon} hold and the iterates belong to the conceptual complete-block extension of Algorithm~\ref{alg:DRAM}.

If \(1\le q_s\le m-r_s\) and $\delta_s=q_s/(m-r_s)$, then for any \(t=s\tau+j\) with $0\le j\le\tau-1$,
\begin{equation} \label{eqn:lem:convert_subspace}
\E\left[\|\nabla F(X_t)\|_*\right]
\le
\frac{1}{\delta_s}\E\left[\|B_s^\top\nabla F(X_t)\|_*\right]+\frac{1+\delta_s}{\delta_s}L_*j\eta.
\end{equation}
Moreover, suppose Assumption~\ref{ass:gradient_filtration} holds with $\sigma=0$. If $0\le q_s\le m-r_s$ with $k_s=r_s+q_s\ge1$, and $\gamma_s=k_s/m$, then
\begin{equation}
\label{eqn:lem:convert_subspace_deterministic}
\E\left[\|\nabla F(X_t)\|_*\right]
\le
\frac{1}{\gamma_s}\E\left[\|B_s^\top\nabla F(X_t)\|_*\right]+\frac{1+\gamma_s}{\gamma_s}L_*j\eta.
\end{equation}
Importantly, the second bound remains valid when $q_s=0$.
\end{lemma}

\begin{proof}
We first prove the stochastic coverage bound, where $q_s>0$.
Let $A:=\nabla F(X_{s\tau})$. Conditional on $(R_s,X_{s\tau})$, the matrix \(A\) is fixed and the only remaining randomness is \(P_s\).
Since $B_s^\top B_s=I$, the matrices $B_s^\top A$ and $B_sB_s^\top A$ have the same nonzero singular values, hence
$\|B_s^\top A\|_*=\|B_sB_s^\top A\|_*$.
By Lemma~\ref{lem:principal-preserving-coverage},
\[\E_{P_s}[B_sB_s^\top\mid R_s,X_{s\tau}]=\Pi_s+\delta_sQ_s.\]
Therefore, by Jensen's inequality,
\begin{align*}
\E_{P_s}\!\left[\|B_s^\top A\|_*\mid R_s,X_{s\tau}\right]
&=\E_{P_s}\!\left[\|B_sB_s^\top A\|_*\mid R_s,X_{s\tau}\right]\\
&\ge\left\|\E_{P_s}[B_sB_s^\top\mid R_s,X_{s\tau}]A\right\|_*\\
&=\|(\Pi_s+\delta_sQ_s)A\|_*\\
&\ge\delta_s\|A\|_*.
\end{align*}
The operator $\Pi_s+\delta_s Q_s$ has singular value $1$ on $\operatorname{Range}(\Pi_s)$ and singular value $\delta_s$ on $\operatorname{Range}(Q_s)$, so $\sigma_{\min}(\Pi_s+\delta_s Q_s)=\delta_s$. Therefore $\|(\Pi_s+\delta_s Q_s)A\|_*\ge \delta_s\|A\|_*$. Taking expectation over $(R_s,X_{s\tau})$ yields $\mathbb{E}\|B_s^\top\nabla F(X_{s\tau})\|_*\ge \delta_s\mathbb{E}\|\nabla F(X_{s\tau})\|_*$.

We next extend the bound from the block start to any \(t=s\tau+j\), where \(0\le j\le \tau-1\). By the triangle inequality and $\|B_s^\top Z\|_*\le\|Z\|_*$,
\begin{align*} \|B_s^\top\nabla F(X_t)\|_* &\ge \|B_s^\top\nabla F(X_{s\tau})\|_* - \|B_s^\top(\nabla F(X_t)-\nabla F(X_{s\tau}))\|_* \\ &\ge \|B_s^\top\nabla F(X_{s\tau})\|_* - \|\nabla F(X_t)-\nabla F(X_{s\tau})\|_* . \end{align*}
Taking expectation and using the block-start bound gives
\[
\mathbb{E}\|B_s^\top\nabla F(X_t)\|_*\ge \delta_s\mathbb{E}\|\nabla F(X_{s\tau})\|_*-\mathbb{E}\|\nabla F(X_t)-\nabla F(X_{s\tau})\|_*.
\]
Moreover, $\|\nabla F(X_{s\tau})\|_*\ge \|\nabla F(X_t)\|_*-\|\nabla F(X_t)-\nabla F(X_{s\tau})\|_*$. Combining the two inequalities gives
\[
\mathbb{E}\|B_s^\top\nabla F(X_t)\|_*\ge \delta_s\mathbb{E}\|\nabla F(X_t)\|_*-(1+\delta_s)\mathbb{E}\|\nabla F(X_t)-\nabla F(X_{s\tau})\|_* .
\]

By nuclear-smoothness, $\|\nabla F(X_t)-\nabla F(X_{s\tau})\|_*\le L_*\|X_t-X_{s\tau}\|_{\op}$. Since \(B_s^\top B_s=I\) and \(\|\,\msign(H_j)\|_{\op}\le1\),
for \(t=s\tau+j\) we have
\begin{align*}
\|X_t-X_{s\tau}\|_{\op}
\le \sum_{f=s\tau}^{t-1} \|X_{f+1}-X_f\|_{\op} = \eta \sum_{f=s\tau}^{t-1} \|B_s\,\msign(H_f)\|_{\op} \le j\eta.
\end{align*}
Therefore $\E\|\nabla F(X_t)-\nabla F(X_{s\tau})\|_*\le L_*j\eta$.
Substituting this estimate into the previous lower bound gives
\[
\E\|B_s^\top\nabla F(X_t)\|_* \ge \delta_s\E\|\nabla F(X_t)\|_* - (1+\delta_s)L_*j\eta .
\]
Rearranging terms proves the stochastic bound~(\ref{eqn:lem:convert_subspace}).

For the deterministic case, Assumption~\ref{ass:gradient_filtration} with $\sigma=0$ implies $G_{s\tau}=\nabla F(X_{s\tau})$ almost surely. Let $A=\nabla F(X_{s\tau})$.

If $r_s=m$, then necessarily $q_s=0$ and $B_s=R_s$ is an
$m\times m$ orthogonal matrix. Hence
\[
\|B_s^\top A\|_*=\|A\|_*=\gamma_s\|A\|_*,
\qquad
\gamma_s=\frac{k_s}{m}=1.
\]
Thus, the desired block-start coverage bound holds directly.

It remains to consider $r_s<m$. Let $A=\nabla F(X_{s\tau})=U\Sigma V^\top$ with $ \sigma_1(A)\ge\cdots\ge\sigma_m(A)\ge0.$
Since $R_s$ is formed by the leading $r_s$ left singular vectors of $G_{s\tau}=A$, we have
\[
\|\Pi_s A\|_*=\sum_{i=1}^{r_s}\sigma_i(A).
\]
Let $\bar\delta_s:=q_s/(m-r_s)$.
When $q_s=0$, $B_sB_s^\top=\Pi_s$.
When $q_s>0$, Lemma~\ref{lem:principal-preserving-coverage} gives $\E_{P_s}[B_sB_s^\top\mid R_s,X_{s\tau}]=\Pi_s+\bar\delta_sQ_s.$
Thus, in both cases,
\begin{equation*}
\E_{P_s}\!\left[\|B_s^\top A\|_*\mid R_s,X_{s\tau}\right]
\ge\|(\Pi_s+\bar\delta_s Q_s)A\|_* =\sum_{i=1}^{r_s}\sigma_i(A)+\bar\delta_s\sum_{i=r_s+1}^{m}\sigma_i(A).
\end{equation*}
Since the singular values are nonincreasing, we obtain $\sum_{i=1}^{r_s}\sigma_i(A)\ge\frac{r_s}{m}\|A\|_*.$ Therefore,
\begin{equation*}
\E_{P_s}\!\left[\|B_s^\top A\|_*\mid R_s,X_{s\tau}\right]
\ge\left[\bar\delta_s+(1-\bar\delta_s)\frac{r_s}{m}\right]\|A\|_* =\frac{r_s+q_s}{m}\|A\|_* =\gamma_s\|A\|_*.
\end{equation*}
Taking expectation yields $\E\|B_s^\top\nabla F(X_{s\tau})\|_*\ge\gamma_s\E\|\nabla F(X_{s\tau})\|_*.$ Repeating the same within-block drift argument used above, with $\gamma_s$ in place of $\delta_s$, gives
\[
\E\|B_s^\top\nabla F(X_t)\|_*\ge\gamma_s\E\|\nabla F(X_t)\|_*-(1+\gamma_s)L_*j\eta,
\]
which is equivalent to~(\ref{eqn:lem:convert_subspace_deterministic}).
\end{proof}

\subsection{Proof of Theorem~\ref{thm:finite_time_bound}} \label{app:proof:DRAM_dynamic_bound}
We use Lemma~\ref{lem:projected_nuclear_coverage} to convert the projected stationarity bound into a full-gradient stationarity bound. Fix a block \(s\) and let \(t=s\tau+j\), where \(0\le j\le \tau-1\). By Lemma~\ref{lem:projected_nuclear_coverage}, \[ \E\left[\|\nabla F(X_t)\|_*\right] \le \frac{1}{\delta_s} \E\left[\|B_s^\top\nabla F(X_t)\|_*\right] + \frac{1+\delta_s}{\delta_s}L_*j\eta . \]
Averaging this inequality over all iterations of the complete-block extension gives
\begin{equation} \label{eqn:thm:DRAM:1}
    \frac1{\widetilde T}\sum_{t=0}^{\widetilde T-1} \E\left[\|\nabla F(X_t)\|_*\right]
\le \frac1{\widetilde T} \sum_{s=0}^{S-1} \frac{1}{\delta_s} \sum_{j=0}^{\tau-1} \E\left[ \|B_s^\top\nabla F(X_{s\tau+j})\|_* \right] + \frac{L_*\eta}{\widetilde T} \sum_{s=0}^{S-1} \frac{1+\delta_s}{\delta_s} \sum_{j=0}^{\tau-1}j.
\end{equation}
Since $\sum_{j=0}^{\tau-1}j=\frac{\tau(\tau-1)}{2}$ and $\widetilde T=S\tau$, the second term becomes
\begin{equation} \label{eqn:thm:DRAM:1.5}
\frac{L_*\eta}{\widetilde T} \sum_{s=0}^{S-1} \frac{1+\delta_s}{\delta_s} \cdot \frac{\tau(\tau-1)}{2} = \frac{L_*\eta(\tau-1)}{2S} \sum_{s=0}^{S-1} \frac{1+\delta_s}{\delta_s}.
\end{equation}

Combining (\ref{eqn:lem:projection_bound}), (\ref{eqn:thm:DRAM:1}) and
(\ref{eqn:thm:DRAM:1.5}) yields
\begin{equation} \label{eqn:thm:DRAM:2}
    \begin{split}
\frac1{\widetilde T}\sum_{t=0}^{\widetilde T-1} \E\left[\|\nabla F(X_t)\|_*\right]
\le& \frac{1}{\eta c_\beta \widetilde T} \sum_{s=0}^{S-1} \frac{\Delta_s}{\delta_s} + \left[ \frac{4(1-\beta)(\tau-1)L_*\eta}{\beta\tau c_\beta} + \frac{L_*\eta}{2c_\beta} \right] \frac1S \sum_{s=0}^{S-1} \frac{1}{\delta_s} \\
&+ \frac{2\sigma}{c_\beta} \sqrt{\frac{\beta}{2-\beta}} \cdot \frac1S \sum_{s=0}^{S-1} \frac{\sqrt{k_s}}{\delta_s} + \frac{L_*\eta(\tau-1)}{2S} \sum_{s=0}^{S-1} \frac{1+\delta_s}{\delta_s}.
    \end{split}
\end{equation}
It remains to control the weighted decrease term. Let  $F_s:=\E[F(X_{s\tau})-F^*]\ge 0$. Then \(\Delta_s=F_s-F_{s+1}\). Since the complementary coverage factor is assumed to be nondecreasing, i.e., $\delta_{s+1}\ge \delta_s$, the sequence \(1/\delta_s\) is nonincreasing. Hence, by summation by parts,
\begin{equation} \label{eqn:thm:DRAM:3}
\sum_{s=0}^{S-1} \frac{\Delta_s}{\delta_s}
= \sum_{s=0}^{S-1} \frac{F_s-F_{s+1}}{\delta_s}
= \frac{F_0}{\delta_0} - \frac{F_S}{\delta_{S-1}} + \sum_{s=1}^{S-1} \left( \frac{1}{\delta_s} - \frac{1}{\delta_{s-1}} \right)F_s
\le \frac{F_0}{\delta_0} = \frac{\Delta}{\delta_0}.
\end{equation}

Substituting (\ref{eqn:thm:DRAM:3}) into (\ref{eqn:thm:DRAM:2}) completes the proof of the stage-dependent full-gradient stationarity bound.

We then consider the deterministic case. Applying Lemmas~\ref{lem:space_muon_one_step_descent} and~\ref{lem:space_muon_tracking} blockwise and setting $\sigma=0$ gives the same per-block estimate as~(\ref{eqn:per_block_Ss}),
\[
\sum_{j=0}^{\tau-1}
\E\|B_s^\top\nabla F(X_{s\tau+j})\|_*
\le
\frac{\Delta_s}{\eta c_\beta}+\frac{4(1-\beta)(\tau-1)L_*\eta}{\beta c_\beta}+\frac{\tau L_*\eta}{2c_\beta}.
\]
Multiplying the above inequality by $1/\gamma_s$, summing over
$s=0,\ldots,S-1$, and using $\widetilde T=S\tau$, we obtain
\[
\begin{aligned}
\frac1{\widetilde T}
\sum_{s=0}^{S-1}\frac{1}{\gamma_s}\sum_{j=0}^{\tau-1}
\E\|B_s^\top\nabla F(X_{s\tau+j})\|_*
\le&
\frac{1}{\eta c_\beta \widetilde T}
\sum_{s=0}^{S-1}\frac{\Delta_s}{\gamma_s}\\
&+\left[\frac{4(1-\beta)(\tau-1)L_*\eta}{\beta\tau c_\beta}+\frac{L_*\eta}{2c_\beta}\right]\frac1S\sum_{s=0}^{S-1}\frac{1}{\gamma_s}.
\end{aligned}
\]
Recall that $F_s=\E[F(X_{s\tau})-F^*]\ge0$ and $\Delta_s=F_s-F_{s+1}$. Since $\gamma_{s+1}\ge\gamma_s$, the sequence
$\{1/\gamma_s\}_{s=0}^{S-1}$ is nonincreasing. Hence, following~(\ref{eqn:thm:DRAM:3}), we obtain $\sum_{s=0}^{S-1}\frac{\Delta_s}{\gamma_s}\leq\frac{\Delta}{\gamma_0}$.
By~(\ref{eqn:lem:convert_subspace_deterministic}) in Lemma~\ref{lem:projected_nuclear_coverage}, for
$t=s\tau+j$,
\[
\E\|\nabla F(X_t)\|_*\le\frac1{\gamma_s}\E\|B_s^\top\nabla F(X_t)\|_*+\frac{1+\gamma_s}{\gamma_s}L_*j\eta.
\]
Averaging over all iterations and using $\sum_{j=0}^{\tau-1}j=\frac{\tau(\tau-1)}{2}$ gives
\begin{equation}
\begin{split}
    \frac1{\widetilde T}\sum_{t=0}^{\widetilde T-1}
\E\|\nabla F(X_t)\|_*
\le&
\frac{\Delta}{\eta c_\beta \widetilde T\gamma_0}+\left[\frac{4(1-\beta)(\tau-1)L_*\eta}{\beta\tau c_\beta}+\frac{L_*\eta}{2c_\beta}\right]\frac1S\sum_{s=0}^{S-1}\frac1{\gamma_s}\\
&+\frac{L_*\eta(\tau-1)}{2S}\sum_{s=0}^{S-1}\frac{1+\gamma_s}{\gamma_s},
\end{split}
\end{equation}
which proves the deterministic statement.

\subsection{Proof of Theorem~\ref{thm:dram_convergence}}
\label{app:proof:DRAM_dynamic_rate}
The actual trajectory and its conceptual complete-block extension coincide through $X_T$. We first bound the extension and then transfer the result to the actual prefix.

Let $A_\delta:=\frac{1}{S}\sum_{s=0}^{S-1}\delta_s^{-1}$, so $\Gamma_\delta=\delta_0^{-1}+A_\delta$. Since $0<\delta_s\le1$, \[\frac{1}{S}\sum_{s=0}^{S-1}\frac{1+\delta_s}{\delta_s}=1+A_\delta\le\Gamma_\delta.\] Moreover, since $k_s\le k_{\max}$, \[\frac{1}{S}\sum_{s=0}^{S-1}\frac{\sqrt{k_s}}{\delta_s}\le\sqrt{k_{\max}}A_\delta\le\sqrt{k_{\max}}\Gamma_\delta.\] The stochastic parameter choice gives $\tau=\lceil3/\beta\rceil$, so $c_\beta\ge1/3$, $\tau-1<3/\beta$, and $(\tau-1)/\tau\le1$. Applying Theorem~\ref{thm:finite_time_bound}, using $A_\delta\le\Gamma_\delta$, $\delta_0^{-1}\le\Gamma_\delta$, $\beta\le1$, and the bound above gives \[\frac{1}{\widetilde T}\sum_{t=0}^{\widetilde T-1}\E\|\nabla F(X_t)\|_*\le3\Gamma_\delta\frac{\Delta}{\eta\widetilde T}+15\Gamma_\delta\frac{L_*\eta}{\beta}+6\Gamma_\delta\sqrt{k_{\max}}\sigma\sqrt{\beta}.\]
Substituting $\eta=\sqrt{\Delta\beta/(5L_*\widetilde T)}$ yields \[\frac{1}{\widetilde T}\sum_{t=0}^{\widetilde T-1}\E\|\nabla F(X_t)\|_*\le6\Gamma_\delta\sqrt{\frac{5L_*\Delta}{\widetilde T\beta}}+6\Gamma_\delta\sqrt{k_{\max}}\sigma\sqrt{\beta}.\]

Let $a:=\sqrt{L_*\Delta/T}$ and $b:=\sqrt{k_{\max}}\sigma$. Since $\widetilde T\ge T$, the preceding bound gives \[\frac{1}{\widetilde T}\sum_{t=0}^{\widetilde T-1}\E\|\nabla F(X_t)\|_*\le6\Gamma_\delta\left(\sqrt{5}\frac{a}{\sqrt{\beta}}+b\sqrt{\beta}\right).\]
By construction, $\beta=\min\{\beta_0,a/b\}$. If $\beta=\beta_0$, then $b\beta_0\le a$, and hence $b\sqrt{\beta_0}\le a/\sqrt{\beta_0}$. If $\beta=a/b$, then $a/\sqrt{\beta}=b\sqrt{\beta}=\sqrt{ab}$.
Therefore, \[\frac{1}{\widetilde T}\sum_{t=0}^{\widetilde T-1}\E\|\nabla F(X_t)\|_*\le6(\sqrt{5}+1)\Gamma_\delta\left(\frac{a}{\sqrt{\beta_0}}+\sqrt{ab}\right).\] Since $\beta_0$ is fixed and $\sqrt{ab}=(k_{\max}L_*\Delta\sigma^2/T)^{1/4}$, \[\frac{1}{\widetilde T}\sum_{t=0}^{\widetilde T-1}\E\|\nabla F(X_t)\|_*=O\!\left(\Gamma_\delta\left[\sqrt{\frac{L_*\Delta}{T}}+\left(\frac{k_{\max}L_*\Delta\sigma^2}{T}\right)^{1/4}\right]\right).\]
Since the stationarity terms are nonnegative, \[\frac{1}{T}\sum_{t=0}^{T-1}\E\|\nabla F(X_t)\|_*\le\frac{\widetilde T}{T}\left[\frac{1}{\widetilde T}\sum_{t=0}^{\widetilde T-1}\E\|\nabla F(X_t)\|_*\right].\] Using $\widetilde T<2T$ proves the stochastic statement of Theorem~\ref{thm:dram_convergence}.

\paragraph{Deterministic case.}
Let $A_\gamma:=\frac1S\sum_{s=0}^{S-1}\frac1{\gamma_s}$, so $\Gamma_\gamma=\gamma_0^{-1}+A_\gamma$. Since $\beta=\beta_0$ is fixed, $\tau=\lceil3/\beta_0\rceil=O(1)$ and $c_\beta\ge1/3$. The deterministic part of Theorem~\ref{thm:finite_time_bound} therefore gives
\[
\frac1{\widetilde T}\sum_{t=0}^{\widetilde T-1}\E\|\nabla F(X_t)\|_*
=O\left(\Gamma_\gamma\left[\frac{\Delta}{\eta\widetilde T}+L_*\eta\right]\right).
\]
Substituting $\eta=\sqrt{\Delta\beta_0/(5L_*\widetilde T)}$ yields
\[
\frac1{\widetilde T}\sum_{t=0}^{\widetilde T-1}\E\|\nabla F(X_t)\|_*
=O\left(\Gamma_\gamma\sqrt{\frac{L_*\Delta}{\widetilde T}}\right).
\]
Applying the same prefix inequality and $T\le\widetilde T<2T$ proves
\[
\frac1T\sum_{t=0}^{T-1}\E\|\nabla F(X_t)\|_*
=O\left(\Gamma_\gamma\sqrt{\frac{L_*\Delta}{T}}\right).
\]

\subsection{Proof of Lemma~\ref{lem:principal-preserving-coverage}}
Let $V\in\mathbb{R}^{m\times(m-r)}$ be an orthonormal basis of $\operatorname{Range}(Q)$, so that $Q=VV^\top$. Since $P$ is uniformly sampled from the $q$-dimensional orthonormal frames in $\operatorname{Range}(Q)$, we can write
\[
P=VU,
\qquad
U\sim\operatorname{Unif}(\mathrm{St}(m-r,q)).
\]
By rotational invariance of the Stiefel distribution,
$\mathbb{E}[UU^\top]$ is invariant under orthogonal conjugation and hence
equals $aI_{m-r}$ for some scalar $a$.
Taking traces gives
\[
q=\mathbb{E}[\operatorname{tr}(U^\top U)]=a(m-r),
\]
so $a=q/(m-r)=\delta$. Therefore,
\[
\mathbb{E}[PP^\top\mid R]=V\mathbb{E}[UU^\top]V^\top=\delta Q.
\]
Since $B=[R\;P]$ and $R^\top P=0$, we obtain $BB^\top=\Pi+PP^\top$, which yields
\[
\mathbb{E}[BB^\top\mid R]
=
\Pi+\delta Q.
\]
Next, since $B^\top B=I$, left multiplication by $B$ preserves the
singular values of $B^\top A$, and hence $\|B^\top A\|_*=\|BB^\top A\|_*.$
By convexity of the nuclear norm and Jensen's inequality,
\begin{align*}
\mathbb{E}\!\left[\|B^\top A\|_*\mid R\right]
=\mathbb{E}\!\left[\|BB^\top A\|_*\mid R\right]
\ge \left\|\mathbb{E}[BB^\top\mid R]A\right\|_*
=\|(\Pi+\delta Q)A\|_*.
\end{align*}

Finally, $\Pi+\delta Q=\delta I+(1-\delta)\Pi$ has smallest singular value $\delta$.
Since $\delta>0$,
\[
\|A\|_*
=
\|(\Pi+\delta Q)^{-1}(\Pi+\delta Q)A\|_*
\le
\|(\Pi+\delta Q)^{-1}\|_{\mathrm{op}}
\|(\Pi+\delta Q)A\|_*
=
\delta^{-1}\|(\Pi+\delta Q)A\|_*.
\]
Therefore, $\|(\Pi+\delta Q)A\|_*\ge\delta\|A\|_*$, which completes the proof.

\subsection{Proof of Corollary~\ref{cor:dram_fixed_allocation}}
For fixed ranks $r_s\equiv r$ and $q_s\equiv q$, we have $k_s\equiv k=r+q$. For $\sigma>0$, choose $k_{\max}=k$. Then $\delta_s\equiv\delta=q/(m-r)$ and hence $\Gamma_\delta=2/\delta$. Direct substitution into the stochastic part of Theorem~\ref{thm:dram_convergence} gives \[\frac{1}{T}\sum_{t=0}^{T-1}\E\|\nabla F(X_t)\|_*=O\!\left(\frac{1}{\delta}\left[\sqrt{\frac{L_*\Delta}{T}}+\left(\frac{L_*\Delta k\sigma^2}{T}\right)^{1/4}\right]\right).\] For $\sigma=0$, $\gamma_s\equiv k/m$, so $\Gamma_\gamma=2m/k$. Substitution into the deterministic part of Theorem~\ref{thm:dram_convergence} gives \[\frac{1}{T}\sum_{t=0}^{T-1}\E\|\nabla F(X_t)\|_*=O\!\left(\frac{m}{k}\sqrt{\frac{L_*\Delta}{T}}\right),\] which proves the claim.

\end{document}